\documentclass[sn-mathphys-num]{sn-jnl}
\usepackage{graphicx}
\usepackage{multirow}
\usepackage{amsmath,amssymb,amsfonts}
\usepackage{amsthm}
\usepackage{mathrsfs}
\usepackage[title]{appendix}
\usepackage[dvipsnames]{xcolor}
\usepackage{textcomp}
\usepackage{manyfoot}
\usepackage{booktabs}
\usepackage{algpseudocode}
\usepackage{listings}
\usepackage{tikz}
\usepackage{longtable}
\usepackage{booktabs}
\usepackage[load-configurations=version-1]{siunitx}

\usepackage{subcaption}
\usepackage{multicol}
\usepackage[ruled,vlined]{algorithm2e}
\usepackage{tabularx}
\usepackage{cleveref}
\theoremstyle{thmstyleone}%
\newtheorem*{remark}{Remark}

\theoremstyle{thmstyletwo}%

\theoremstyle{thmstylethree}%
\newtheorem{definition}{Definition}%

\begin{document}

\title[Sample-Efficient RL with Symmetry-Guided Demos for Robotic Manipulation]{Sample-Efficient Reinforcement Learning with Symmetry-Guided Demonstrations for Robotic Manipulation}


\author[1]{\fnm{Amir Mehdi} \sur{Soufi Enayati}}\email{amsoufi@uvic.ca}
\author[2]{\fnm{Zengjie} \sur{Zhang}}\email{z.zhang3@tue.nl}
\author[1]{\fnm{Kashish} \sur{Gupta}}\email{kashishg@uvic.ca}
\author*[1]{\fnm{Homayoun} \sur{Najjaran}}\email{najjaran@uvic.ca}

\affil[1]{\orgdiv{Department of Mechanical Engineering}, \orgname{University of Victoria}, \orgaddress{\city{Victoria}, \state{BC}, \country{Canada}}}

\affil[2]{\orgdiv{Department of Electrical Engineering}, \orgname{Eindhoven University of Technology}, \orgaddress{\city{Eindhoven}, \country{Netherlands}}}

\abstract{
{Reinforcement learning (RL) suffers from low sample efficiency, particularly in high-dimensional continuous state-action spaces of complex robotic manipulation tasks. RL performance can improve by leveraging prior knowledge, even when demonstrations are limited and collected from simplified environments.
To address this, we define General Abstract Symmetry (GAS) for aggregating demonstrations from symmetrical abstract partitions of the robot environment. We introduce Demo-EASE, a novel training framework using a dual-buffer architecture that stores both demonstrations and RL-generated experiences. Demo-EASE improves sample efficiency through symmetry-guided demonstrations and behavior cloning, enabling strong initialization and balanced exploration-exploitation.
Demo-EASE is compatible with both on-policy and off-policy RL algorithms, supporting various training regimes. We evaluate our framework in three simulation experiments using a Kinova\textregistered~Gen3 robot with joint-space control within PyBullet. Our results show that Demo-EASE significantly accelerates convergence and improves final performance compared to standard RL baselines, demonstrating its potential for efficient real-world robotic manipulation learning.}
}


\maketitle

\section{Introduction}\label{sec:intro}
A major limitation of RL preventing its wide applications to robotic systems is its low sampling efficiency~\cite{dulac2021challenges, buckman2018sample}. This is mainly due to the high dimensional continuous state and action spaces of the motion planning problem to be solved for complex manipulation tasks~\cite{SOUFIENAYATI2022381}. As a result, a large amount of data is required for an RL agent to achieve a decent training performance, although only a small portion benefits the policy improvement. Thus, training an RL agent for robot manipulation tasks is usually expensive and inefficient. Several techniques have been used to tackle this challenge. For example, compact state representation can be used to shrink the search space~\cite{ren2021experimental, ejaz2020vision}. Hierarchical reinforcement learning (HRL)~\cite{li2020hrl4in,hou2020data,yang2021hierarchical} can be used to create simpler sub-problems based on low-dimensional abstractions. Similarly, curriculum for efficient sweeping over the state space~\cite{gupta2022extending}, or heuristic dynamic modeling~\cite{stulp2012reinforcement} have been proposed to reduce the task dimensionality. These methods either reduce unnecessary exploration or remap the policy search space to a lower-dimensional domain.

Another two types of heuristics commonly used to improve the efficiency of RL are \textit{symmetry} and \textit{demonstrations}. \textit{Symmetry} helps create a shared abstract model for different domains of the environment using a common feature set~\cite{mahajan2017symmetry}. With a well-defined symmetry, the features, namely state, and actions for the case of RL, can be mapped among different symmetric domains without additional interaction. This minimizes the model complexity and decisively improves the sampling efficiency. In previous work~\cite{Gupta_2021,gupta2022icra}, the symmetry of a complex environment has been used to facilitate efficient sampling of RL policy training, forming a general learning framework named Exploitation of Abstract Symmetry of Environments (EASE). EASE uses RL training for both abstract and global environments. Thus, it suffers from RL sample inefficiency but to a lesser degree. This drawback can hopefully be mitigated by utilizing \textit{demonstrations}, a set of reference trajectory clips of the system states performed by experts~\cite{hersch2008dynamical}. Unlike in programming by demonstration (PbD)~\cite{argall2009survey}, demonstration-based RL for robotic manipulation does not aim to imitate sub-optimal human motion since fanatical imitation may lead to over-fitting and lack of generalizability. Thus, behavior cloning is needed to exploit demonstrations without causing over-fitting~\cite{zhang2024using}. Leveraging the data aggregation technology~\cite{calderon2024deep}, symmetry can be used to improve the exploitation of demonstrations by mapping the data among symmetric domains, thus promoting the efficiency of the learning process of an RL agent.

{Additionally, there have been efforts in improving sampling efficiency in robot learning by improving action representation and architectural inductive biases. The Semantic-Geometric Representation (SGR)\cite{zhang2023sgr} introduced visual fusion with pure behavior cloning to achieve strong visuomotor control. Its successor, SGRv2\cite{zhang2024sgrv2}, extended SGR by incorporating a strong inductive bias, i.e., action-locality, assuming actions are predominantly influenced by the target object. Both of these methods rely on demonstrations and follow an imitation learning paradigm for policy discovery. Hence, in this case, sample efficiency translates to requiring the minimal number of demonstrations used for policy learning. From a different perspective on sample efficiency, architectural symmetry has been embedded in model design in \textit{equivariant} RL to exploit spatial invariances. For instance, SE(2)-equivariant Q-learning~\cite{wang22j} was proposed in fully observable tasks, later extended to partially observable settings~\cite{nguyen23a} via symmetry-aware actor-critic networks. However, these approaches use exact symmetry in the network design and do not explore learning from demonstrations or knowledge transfer. In contrast, our approach explicitly uses demonstrations and domain-level symmetry in the environment for improved data aggregation and efficient RL, while maintaining model-agnostic compatibility.
}

In this paper, we propose a novel RL training framework for robot manipulation tasks, named Demo-EASE, facilitated by symmetry-guided demonstrations. The demonstrations are stored in a demo buffer, which, together with the experience buffer, forms a dual-memory learning scheme.
{The additional computational cost of this dual-memory architecture is minimal compared to the sample efficiency gains. The main bottleneck in RL for robotics is typically the cost of environment interactions and failed convergence, rather than memory usage or data preprocessing.}

A formally defined General Abstract Symmetry (GAS) allows for aggregation of demonstrations in symmetric partitions of the environment. 
A behavior cloning (BC) loss function is designed to update the actor-critic agent with the dual-memory structure~\cite{nair2018overcoming}. In this way, Demo-EASE extends the previous EASE framework originally designed for discrete spaces and off-policy agents to on-policy agents with continuous spaces, achieving improved exploitation efficiency and promoting training speed. 
A comparison study has been conducted between the proposed method and two RL benchmarks, namely Deep Deterministic Policy Gradient (DDPG)~\cite{lillicrap2015continuous}, as the base off-policy method and Proximal Policy Optimization (PPO)~\cite{schulman2017proximal} as the base on-policy method, showcasing that the Demo-EASE framework can effectively improve the training speed and score of an RL agent for robot manipulation.

{DDPG is an off-policy RL method with a deterministic actor, and PPO is an on-policy RL method with a stochastic actor. This choice covers two major classes of actor–critic RL methods, differing in memory structure, action modeling, and value function estimation. Notably, these algorithms remain standard baselines in recent works on robotic manipulation~\cite{wang2023dexterous,10343126}, making them thoroughly and practically representative. The modular design of Demo-EASE, relying only on the availability of a value function and an actor, ensures that it can be seamlessly extended to other actor-critic methods such as SAC, TD3, A2C/A3C, and TRPO. The formulation of behavior cloning loss and value-based masking strategies further supports this extensibility, as they only require critic outputs and do not depend on the policy update scheme.}

The rest of the paper is organized as follows. Section \ref{sec:form} formulates the problem to be investigated in this paper. The technical details of the proposed method are presented in Section \ref{sec:method}. The implementation results are discussed in Section \ref{sec:results}. Finally, Section \ref{sec:conclusion} concludes the paper. Additional information about the algorithm and experiment setup can be found in Sections \ref{sec:method_rl} and \ref{sec:exp}.

\textit{Notation and units:} We use $\mathbb{R}$, $\mathbb{R}^+$, $\mathbb{R}^n$, $\mathbb{N}^+$ to represent the sets of real numbers, positive real numbers, $n$-dimensional real vectors, and positive integers, respectively. $\lvert \cdot \rvert$ is the absolute value of a scalar and $\|\cdot\|$ stands for the $2$-norm of any real vector. All angles are in radian and all lengths are in meters, if not specified.

\section{Problem Statement}\label{sec:form}

Three essential robot tasks are studied in this paper as representatives of general robot manipulation tasks, namely a point-to-point reaching (P2P) task, a P2P task with obstacle avoidance (P2P-O), and a pick-and-place (P\&P) scenario. These three tasks are the most essential primitives to form more complicated manipulation tasks. Based on these primitives, we describe a general robot motion planning task as a Markov decision process (MDP) and solve it using RL.

\subsection{Robot Manipulation as a MDP}

We consider an MDP $\mathcal{M} := (\mathcal{S},\mathcal{A}, f, R, \gamma)$, where $\mathcal{S}$ and $\mathcal{A}$ are the state and action spaces of the robotic system, $f: s' \sim p(\cdot|s, a)$ represents the state transition of $\mathcal{M}$, with $s, s' \in \mathcal{S}$ and $a \in \mathcal{A}$, $R: \mathcal{S} \times \mathcal{A} \rightarrow \mathbb{R}$ is the task-specific reward function, and $\gamma \in (0,1]$ is the discounting factor. The target of the MDP problem is to solve the optimal policy $\pi: a \sim \pi(\cdot|s)$ such that the following accumulated reward is maximized,
\begin{equation}\label{eq:optideter}
\textstyle J(\pi) := \mathbb{E}_{\tau \sim \pi} \!\left( \sum_{t=0}^{T} \gamma^{t} R\!\left(s_t,a_t \right) \right),
\end{equation}
where $T$ is the length of the robot trajectory $\tau(\pi):=\{s_0,a_0,\cdots,s_{T-1},a_{T-1},s_{T}\}$, and $s_i \in \mathcal{S}$ and $a_i \in \mathcal{A}$ are respectively the observed state and action. 

For an $n$-degree-of-freedom ($n$-DOF) robot manipulator, we define the action of all three tasks as the vector of reference joint velocities, i.e., $a_t = \dot{q}_t^{\mathrm{r}} \in \mathbb{R}^n$, where $q_t^{\mathrm{r}} \in \mathbb{R}^n$ is the joint angles of the robot with the subscript $t$ indicating the time and the superscript $\mathrm{r}$ referring \textit{reference trajectory}. 
The state of the P2P task is defined as
\begin{equation}\label{eq:st_p2p}
    s_t = \left[ \, q_t, \, \sin(q_t), \, \cos(q_t), \, \dot{q}_t, \, P_g, \, e_t  \, \right],
\end{equation}
where $q_t, \dot{q}_t \in \mathbb{R}^n$ are respectively the actual joint angles and joint velocities at time $t$, $P_g \in \mathbb{R}^3$ is the Cartesian position of the goal, and $e_t = P_g - P_t^e$ is the vector distance between the goal position $P_g$ and the robot end-effector position $P_t^e \in \mathbb{R}^3$. The state vector of the P2P-O and P\&P tasks are defined similarly to \eqref{eq:st_p2p} with an addition of the object position $P_o \in \mathbb{R}^3$. Here, we assume the object is a certain cube and $P_o$ is its geometrical center.

\subsection{Solving MDP Using RL}\label{sec:qvl}

RL can solve the optimal policy $\pi^*$ of an MDP $\mathcal{M}$ by exploiting the data generated during the interactions between a motion planning agent and the environment. The critical technical point of RL is to use the data to approximate the optimal value function $Q(s, a):=\max_{\pi}Q^{\pi}(s, a)$, where $Q^{\pi}(s, a)$ is the policy value function associated to $\pi$, defined as
\begin{equation}
\textstyle    Q^{\pi}(s,a):= \mathbb{E}_{\tau \sim \pi} \left( \sum_{t=0}^{T} \gamma^{t} R\!\left(s_t,a_t \right)|s_0=s,a_0=a\right).
\end{equation}
Then, the optimal policy can be obtained as $\pi^*(s):= \arg \max_{a \in \mathcal{A}} Q(s,a)$. The training stage of an RL agent renders a bootstrapping process, where the policy $\pi$ is evaluated (estimating $Q^{\pi}(s, a)$ using $\pi$) and improved (updating $\pi$ using $Q^{\pi}(s, a)$) iteratively. This is the main reason for the low exploitation of data. 

An RL model used to solve the optimal policy $\pi^*:= \arg \max_{\pi}J(\pi)$ of the MDP $\mathcal{M}$ is also referred to as an RL \textit{agent}.
According to the way that the policy is estimated and improved, RL methods can be categorized as \textit{on-policy} and \textit{off-policy}~\cite{pmlr-v119-fedus20a}. The former uses the policy's value to improve itself, while the latter allows for improving the policy using the value of another policy.

\subsection{Facilitate RL training using Demonstrations}

Demonstration refers to the data collected when an expert agent is deployed to interact with the environment. A demonstration episode is a trajectory of the MDP $\mathcal{M}$ \[\tau(\pi^{\mathrm{d}}) := \left\{s_0, a_0, s_1, a_1, \cdots, s_{T-1}, a_{T-1}, s_{T} \right\}, T \in \mathbb{N}^+,\]
where $\pi^{\mathrm{d}}$ is the policy used to generate the demonstration. The demonstration policy can be generated from previous data~\cite{nair2018overcoming}, or explicit teaching~\cite{taylor2011integrating}, or an imperfect controller~\cite{gao2018reinforcement}. Demonstration serves as prior knowledge to guide the agent through the initial learning stage. It is believed that the demonstrated policy $\pi^{\mathrm{d}}$ is closer to the optimal policy $\pi^*$ than the initial policies, making the demonstration data more efficient in improving the current policy. However, the demonstration can cause a local-optima problem in the later stages of agent training~\cite{s21041278}. Therefore, the agent should use the demonstration knowledge in the early learning stage and start to balance between the demonstration and its own experience when it becomes mature.

There is a difference between off-policy and on-policy agents in exploiting demonstrations. The former can utilize demonstrations generated by any demonstration policies, while the latter has to update demonstrations online during agent training.

\section{The Demo-EASE Framework}\label{sec:method}

In this section, we present the proposed Demo-EASE training framework for RL agents. Section \ref{sec:envs} introduces the General Abstract Symmetry (GAS), the fundamental concept of the Demo-EASE framework. Section \ref{sec:demo} explains the generation and aggregation of symmetric demonstrations using GAS. Then, the training framework and algorithms of Demo-EASE are introduced in Section \ref{sec:method_rl}, with Section \ref{sec:loss} giving the critical definition of training loss functions.

\subsection{General Abstract Symmetry (GAS)}\label{sec:envs}

The main technical point of this paper is to improve the sampling efficiency of RL algorithms using symmetry-guided demonstrations. Conventionally, symmetry has been used as a task-specific notion, lacking a clear definition~\cite{Gupta_2021}. In this section, we provide a precise definition of symmetry associated with a general MDP.\\

\begin{definition}\label{def:symmetry}
For an MDP $\mathcal{M}:=(\mathcal{S}, \mathcal{A}, f, R, \gamma)$ as defined in Sec.~\ref{sec:form}, the sub-domains $\mathcal{S}_1, \mathcal{S}_2 \subset \mathcal{S}$ are \textit{symmetric} to each other, or $\mathcal{S}_1 \rightleftharpoons \mathcal{S}_2$, if there exist reversible mappings $\phi: \mathcal{S}_2 \rightarrow \mathcal{S}_1$ and $\psi: \mathcal{A} \rightarrow \mathcal{A}$, such that
1). $Q(s, a) = Q(\phi(s),\psi(a))$ holds for any $s \in \mathcal{S}_2$ and $a \in \mathcal{A}$, and
2). $Q(s, a) = Q(\phi^{-1}(s),\psi^{-1}(a))$ holds for any $s \in \mathcal{S}_1$ and $a \in \mathcal{A}$,
where $Q(s,a)$ is the optimal value function given in Sec.~\ref{sec:qvl}.\\
\end{definition}

{The invariance of the optimal value function under the symmetry mappings builds on the dynamic invariance under symmetry. The invariance of the state transition function, $f: s' \sim p(\cdot|s, a)$, under the symmetry mappings $\phi$ and $\psi$ implies that applying these mappings before or after a state transition yields consistent outcomes~\cite{zinkevich2001symmetry}. Namely, the mapping of the next state $\phi(s')$ should correspond to the result of executing the mapped action $\psi(a)$ from the mapped state $\phi(s)$; i.e.,
\begin{equation*}
     \phi(f(s,a)) = f(\phi(s), \psi(a)) \Leftrightarrow p(\phi(s') \mid \phi(s), \psi(a)) = p(s' \mid s, a)
\end{equation*}
for any $s \in \mathcal{S}_2$ and any $a \in \mathcal{A}$. Similarly, the reward function is expected to satisfy
\begin{equation*}
R(s, a) = R(\phi(s), \psi(a)).
\end{equation*}
These conditions ensure that the dynamics are consistent across symmetric subdomains of the MDP. They also justify the symmetry relation in Definition~\ref{def:symmetry}, and they lead to invariance of the optimal value function $Q(s, a) = Q(\phi(s), \psi(a))$. \\

\begin{remark}[Invariance of Dynamics]
Let $\mathcal{S}_1 \rightleftharpoons \mathcal{S}_2$ with reversible symmetry mappings $\phi: \mathcal{S}_2 \rightarrow \mathcal{S}_1$ and $\psi: \mathcal{A} \rightarrow \mathcal{A}$ as in Definition~\ref{def:symmetry}. For any $s \in \mathcal{S}_2$ and $a \in \mathcal{A}$:
\begin{align*}
    \phi(f(s,a)) &= f(\phi(s), \psi(a)), \\
    R(\phi(s), \psi(a)) &= R(s, a).
\end{align*}
Under these conditions, the optimal value function satisfies the invariance stated in Definition~\ref{def:symmetry}
\begin{equation*}
Q(s, a) = Q(\phi(s), \psi(a)) \quad \text{and} \quad Q(s, a) = Q(\phi^{-1}(s), \psi^{-1}(a)).
\end{equation*}
\end{remark}

\begin{proof}
Assuming the transition and reward functions of the MDP $\mathcal{M}$ are invariant under the mappings $\phi$ and $\psi$ for any $s \in \mathcal{S}2$ and any $a \in \mathcal{A}$
\begin{align*}
    \phi(f(s,a)) = f(\phi(s), \psi(a)) &\Leftrightarrow p(\phi(s') \mid \phi(s), \psi(a)) = p(s' \mid s, a),\\
    R(\phi(s), \psi(a)) &= R(s, a)
\end{align*}
Accordingly, the Bellman optimality equation for the optimal value function $Q$ under these mappings becomes
\begin{align*}
    Q(\phi(s), \psi(a)) &=  \sum_{\phi(s')} p(\phi(s') \mid \phi(s), \psi(a)) \left[ R(\phi(s), \psi(a)) + \gamma \max_{a'} Q(\phi(s'), a') \right] \\
    &= \sum_{s'} p(s' \mid s, a) \left[ R(s, a) + \gamma \max_{a'} Q(s', a') \right] \\
    &= Q(s, a).
\end{align*}
By a similar argument, one can verify that $Q(\phi^{-1}(s), \psi^{-1}(a)) = Q(s, a)$. This satisfies the symmetry condition stated in Definition~\ref{def:symmetry}.
\end{proof}

}

\begin{definition}[General Abstract Symmetry]\label{def:sps}
For the MDP $\mathcal{M}$, we refer to the domain $\mathcal{S}$ as \textit{symmetrically partitionable}, if $\exists\, \mathcal{S}_1$, $\mathcal{S}_2$, $\cdots$, $\mathcal{S}_{K}$ $\subset$ $\mathcal{S}$, $K \in \mathbb{N}^+$, such that $\mathcal{S}_1 \cup \mathcal{S}_2 \cup \cdots \cup \mathcal{S}_{K} = \mathcal{S}$, and $S_{i} \rightleftharpoons \mathcal{S}_{j}$ and $S_{i} \cap \mathcal{S}_{j} = \varnothing$ for all $i,j$$\,=\,$$1$, $2$, $\cdots$, $K$ and $i \neq j$. Then, the domain set $\{\mathcal{S}_{1}, \mathcal{S}_{2}, \cdots, \mathcal{S}_{K}\}$ is subject to a GAS. \\
\end{definition}

Definition \ref{def:sps} introduces a special type of MDP, of which the state space can be split into several mutually exclusive and symmetric partitions. On such MDP, a special type of symmetry relation named GAS is defined. \textit{Symmetry} indicates that any sample $(s,a)$ with $s \in \mathcal{S}$ and $a \in \mathcal{A}$ can find itself located in one and only one partition, and can find its symmetric counterparts $(\phi(s), \psi(a))$ on any other symmetric partitions. This means that all data sampled from these symmetric partitions can be exploited similarly due to having a common value, i.e., $Q(s, a) = Q(\phi(s), \psi(a))$, addressing the possibility of improving the sampling efficiency of an RL agent via data aggregation from different partitions. \textit{Abstract} means that the domain set $\{\mathcal{S}_1, \mathcal{S}_2, \cdots, \mathcal{S}_K\}$ forms an abstraction of the MDP $\mathcal{M}$. A hierarchical RL framework can be designed with its lower level controlling the state transition within each partition and its higher level in charge of switching between different partitions. Finally, \textit{General} implies that the definitions above hold for any general MDPs.

\begin{figure}[htbp]
    \centering
    \begin{tikzpicture} \node[] () at (0,0){\includegraphics[width=0.5\linewidth,clip]{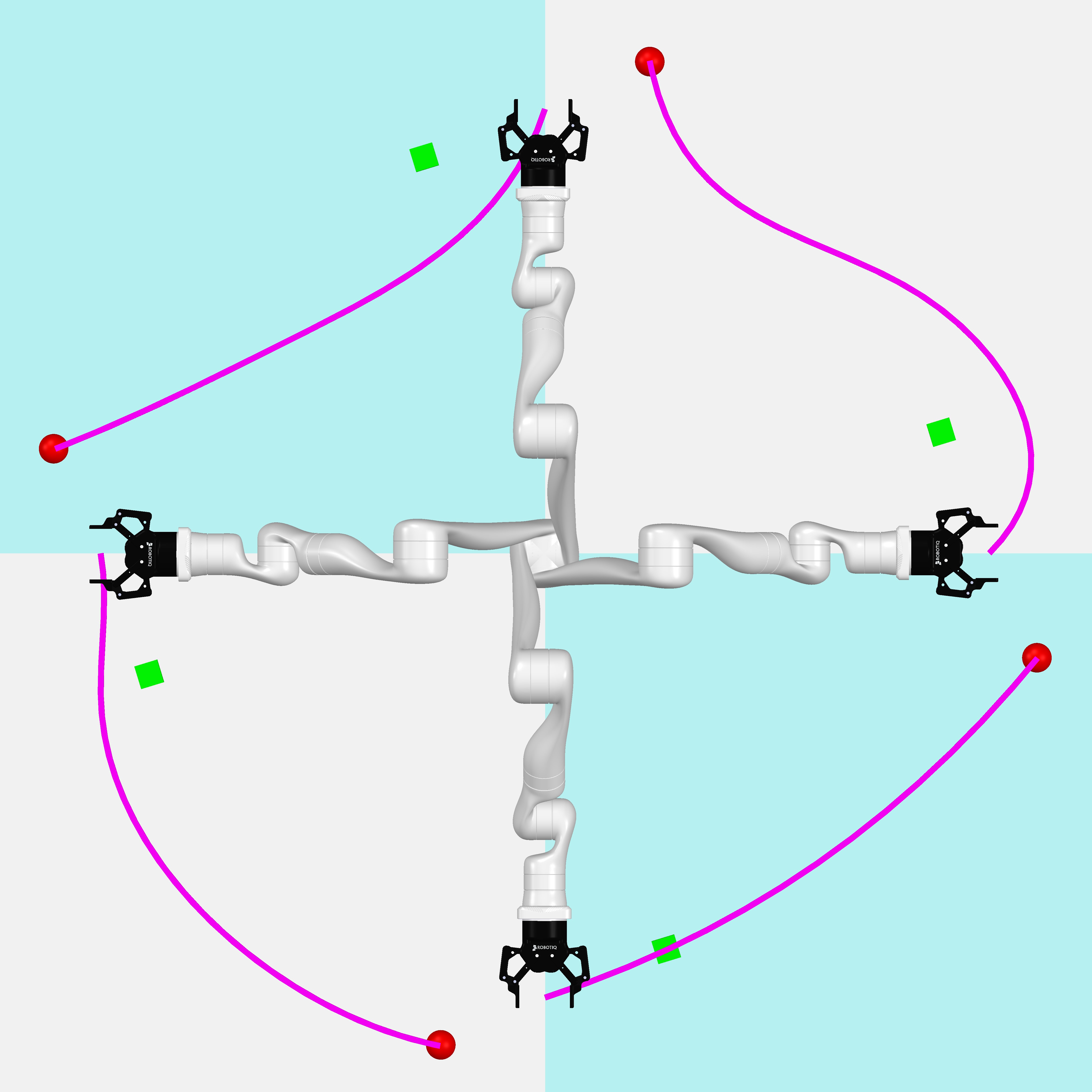}};
    \node[] () at (3,3) {$\mathcal{S}_1$};
    \node[] () at (-3,3) {$\mathcal{S}_2$};
    \node[] () at (-3,-3) {$\mathcal{S}_3$};
    \node[] () at (3,-3) {$\mathcal{S}_4$};
    \end{tikzpicture}
    \caption{Symmetric partitions of the 2D workspace of robot manipulation}
    \label{fig:quad}
\end{figure}

Since the true Q-value $Q(s, a)$ is difficult to obtain, it is very challenging to determine whether an MDP is symmetrically partitionable and to solve all its symmetric partitions purely using the two definitions above. In practice, determining symmetric local environments from a global environment still relies on empirical methods, such as utilizing the geometric symmetry of a robot and its operation space. The previous work~\cite{gupta2022icra} has demonstrated the geometric symmetry of a discrete-state environment using a puzzle example. Now, we explain GAS using a robot manipulation example. As shown in Figure \ref{fig:quad}, a robot manipulator is based in the geometry center of a workspace $\mathcal{S}$. For clarity, we only show the 2D top-to-bottom view of the workspace. We split the entire workspace $\mathcal{S}$ into four partitions, $\mathcal{S}_1:=\{ (\theta, \rho) \mid 0 \leq \theta < {\pi}/{2},\, 0< \rho \leq L \}$, $\mathcal{S}_2:=\{ (\theta, \rho) \mid {\pi}/{2} \leq \theta <  \pi,\, 0< \rho \leq L \}$, $\mathcal{S}_3:=\{ (\theta, \rho) \mid \pi \leq \theta < {3\pi}/{2} ,\, 0< \rho \leq L \}$, and $\mathcal{S}_4:=\{ (\theta, \rho) \mid {3\pi}/{2} \leq \theta < 2 \pi,\, 0< \rho \leq L\}$, where $(\theta, \rho)$ denote the polar coordinates and $L$ is the reaching range of the manipulator.

Now, we justify the GAS among the four partitions. Without losing generality, we take $\mathcal{S}_1$ and $\mathcal{S}_2$ as an example and ignore the stochasticity of the state transitions. Subject to the same control policy, it is straightforward that the resulting robot trajectories also have the same shape, other than that they are relatively rotated to the base of the robot. In fact, for any robot state $s \in \mathcal{S}_1$ and action $a \in \mathcal{A}$ such that the successive state $s' \in \mathcal{S}_1$, there always exist two invertible mappings $\phi: \mathcal{S}_1 \rightarrow \mathcal{S}_2$ and $\psi: \mathcal{A} \rightarrow \mathcal{A}$, where $\psi(a) = a$ and
\begin{equation}\label{eq:st_p2p_phi}
    \phi(s_t) = \left[ \, q_t^{\phi}, \, \sin(q_t^{\phi}), \, \cos(q_t^{\phi}), \, \dot{q}_t, \, P_g, \, e_t^{\phi}  \, \right],
\end{equation}
where $q_t^{\phi}$ adds ${\pi}/{2}$ to the robot base joint angle compared to $q_t$, such that $\phi(s) \in \mathcal{S}_1$ and $\phi(s)' = \phi(s') \in \mathcal{S}_1$, and $e_t^{\phi} = P_g-(P_t^e)^{\phi}$ is the reaching error, where $(P_t^e)^{\phi}$ is the transformed end-effector position from $P_t^e$ with rotation angle $\pi/2$ relative to the origin. As a result, the successive trajectories in the two partitions have the same shape and the same value. This holds for all samples in $\mathcal{S}_1$ and vice versa. Thus, according to Definition \ref{def:symmetry}, we know that $\mathcal{S}_1$ and $\mathcal{S}_2$ are symmetric. Such a property can easily be verified for other partitions, which leads to the fact that the four partitions are symmetric. According to Definition \ref{def:sps}, the workspace $\mathcal{S}$ is symmetrically partitionable and $\{\mathcal{S}_i\}$, $i=1,2,3,4$, are subject to a GAS relation. 

In this paper, we refer to a symmetrically partitionable workspace as the \textit{global environment}, and any of its symmetric partitions as a \textit{local environment}.

\subsection{Demonstration Aggregation Using GAS}\label{sec:demo}

In this section, we interpret how to utilize GAS to facilitate the aggregation of demonstration data. We generate the demonstration data using a PID controller due to simplicity and robustness. For any time instant $t=0,1,\cdots,T$ and state $s_t$, a PID action $a_t^{\mathrm{PID}} \sim \pi^{\mathrm{d}}(s_t)$ is defined as
\begin{equation}\label{eq:pid}
\textstyle a_t^{\mathrm{PID}} = K_{\mathrm{P}} \tilde{q}_t + K_{\mathrm{I}} \int_0^t \tilde{q}_{\tau} \mathrm{d}\tau + K_{\mathrm{D}} \dot{\tilde{q}}_t,
\end{equation}
where $\tilde{q}_t = q_t^{\mathrm{sp}} - q_t$, $q_t^{\mathrm{st}} \in \mathbb{R}^n$ is the desired joint angles obtained by calculating the inverse kinematics of the Cartesian set points, and $q_t$ is the current joint angles of the robot. For the P2P and P\&P tasks, $q_t^{\mathrm{sp}}$ is the inverse kinematics of the goal position $P_g$. For the P2P-O task, heuristic midway points are added to $q_t^{\mathrm{sp}}$ such that the robot moves away from the obstacle. $K_{\mathrm{P}}$, $K_{\mathrm{I}}$, and $K_{\mathrm{D}}$ are respectively the proportional gain, the integral gain, and the derivative gain of the PID controller. With the state $s_t$ and action $a_t^{\mathrm{PID}}$, the successive state $s'_t$ is observed and the instant reward $r_t = R(s_t, a_t^{\mathrm{PID}})$ is calculated. Although a PID controller has been used to generate the demonstration data, it can be substituted with other planning methods or human motion. 

For a manipulation task in a GAS environment, the demonstration only needs to be generated in one local environment. According to the definition of GAS, any trajectory $\tau(\pi^{\mathrm{d}}):=\{s_0,a_0, \cdots,s_{T-1}, a_{T-1}, s_T\}$ within this local environment can find its symmetric counterpart $\tau(\pi^{\mathrm{d}'}):=\{\phi(s_0),\psi(a_0), \cdots,\phi(s_{T-1}), \psi(a_{T-1}), \phi(s_T)\}$ in any other symmetric local environment, ensuring $Q(s_t, a_t) = Q(\phi(s_t), \psi(a_t))$ for all $t=0,1,\cdots,T-1$, given that the invertible domain mappings $\phi$ and $\psi$ exist. This means that, for any sample $(s_t, a_t)$ collected on a local environment, another $K-1$ samples $(\phi(s_t), \psi(a_t))$ can be mapped without real demonstrations. Choosing the four quadrants for partitioning is arbitrary and a design choice based on the size of the workspace.
This also makes it possible to aggregate the data in $\tau(\pi^{\mathrm{d}})$ and $\tau(\pi^{\mathrm{d}'})$, even though that the demonstration policies $\pi^{\mathrm{d}}$ and $\pi^{\mathrm{d}'}$ may be different. Such a property is especially useful for on-policy agents, which require specific policies for policy improvement.

Let us take an off-policy RL agent as an example to address how aggregation is performed, where the demonstration is buffered in the form of $\mathcal{D}_t = \left\{s_t, a^{\mathrm{PID}}_t, s'_t, r_t, \zeta_t \right\}$, where $\zeta_t \in \{0,1\}$ is a binary termination flag enabled by success, timeout, collision, or exiting the local environment. Then, a duplicated demonstration $\mathcal{D}_t' = \{\phi(s_t), \psi(a^{\mathrm{PID}}), \phi(s_t)', r_t, \zeta_t\}$ can be aggregated to the demonstration buffer, where $\phi(s_t)' \sim p(\cdot|\phi(s_t),\psi(a^{\mathrm{PID}}))$ is the successive state observed based on state $\phi(s_t)$ and action $\psi(a^{\mathrm{PID}}))$. If the state transition $f$ is assumed to be Gaussian and $\phi$ is a linear, $\phi(s_t)'$ can be estimated by $\phi(s_t)' = \phi(s_t')$ without real observation.

GAS-based demonstration aggregation can improve the sampling efficiency of RL agent training. Real sampling only needs to be performed on one symmetric local environment, which can generate times of demonstrations without real sampling. In this way, GAS can leverage the exploitation of a large-scale demonstration set with little real sampling.

\subsection{Design and Implementation of Demo-EASE}
\label{sec:method_rl}

The overall structures of the proposed Demo-EASE framework for off-policy and on-policy agents are illustrated in Figures~\ref{fig:arch_ddpg} and \ref{fig:arch_ppo}, respectively. Inspired by the previous EASE framework~\cite{gupta2022icra}, Demo-EASE retains a dual-buffer structure with an original experience buffer and a demo buffer. The original buffer collects the agent's interactions with the global environment, while the demo buffer stores demonstration samples collected in the local environments, i.e., the symmetric partitions.

Unlike the conventional EASE framework \cite{gupta2022icra}, Demo-EASE extends to continuous state and action spaces. Moreover, for the off-policy case, instead of generating demonstrations using a locally trained DDPG agent as in prior work, we focus on transferring knowledge from a pre-existing but sub-optimal controller. This choice enhances practicality by leveraging existing control systems. For the on-policy case, the training process is a complete direct extension using a similar dual-buffer structure. Although the proposed implementation is demonstrated with only two RL algorithms, this method can be extended to other actor-critic RL methods equipped with online or offline experience replay buffers.

To further clarify the versatility of Demo-EASE, we highlight the key implementation differences between its off-policy and on-policy versions. In off-policy algorithms like DDPG, demonstration data can be inserted into the replay buffer at initialization, allowing the agent to reuse them throughout training. In contrast, on-policy methods such as PPO use temporary online buffers, and demonstration samples must be accumulated gradually during early rollouts.

Note that there is a difference between the off-policy and on-policy algorithms in how the demonstrations are exploited. The offline demo buffer structure is similarly shared between the off-policy and on-policy Demo-EASE. But for the latter case, we include demonstrations in the online buffer solely during early iterations, i.e., for the first or a considerably small number of experiences of the early epochs, until filling the demo buffer to its nominal capacity. This allows the agent to learn directly from demo interactions only in the early epochs, while the influence of behavior cloning regulation is present throughout the training as a result of the offline demo buffer.

In addition, off-policy algorithms use action-value functions $Q(s,a)$ for the critic, where on-policy algorithms employ state-value functions $V(s)$. This distinction determines different masking rules applied during behavior cloning corresponding to Equations~\ref{eq:mask_q} and~\ref{eq:mask_v}, respectively. Demo-EASE can also incorporate both deterministic and stochastic actors for behavior cloning loss calculation. With a deterministic actor, as in the DDPG-based variant, the predicted action is the direct output of the actor, and with a stochastic actor, as in the PPO-based variant, the predicted action is the mean of the action distribution. These differences are handled with minimal architectural changes, highlighting the modularity of Demo-EASE. They also prove the adaptability and suitability of our framework across other actor–critic RL algorithms.

\begin{figure}[htbp]
\centering
    \begin{subfigure}[b]{\textwidth}
        \centering
        \includegraphics[width=0.96\textwidth]{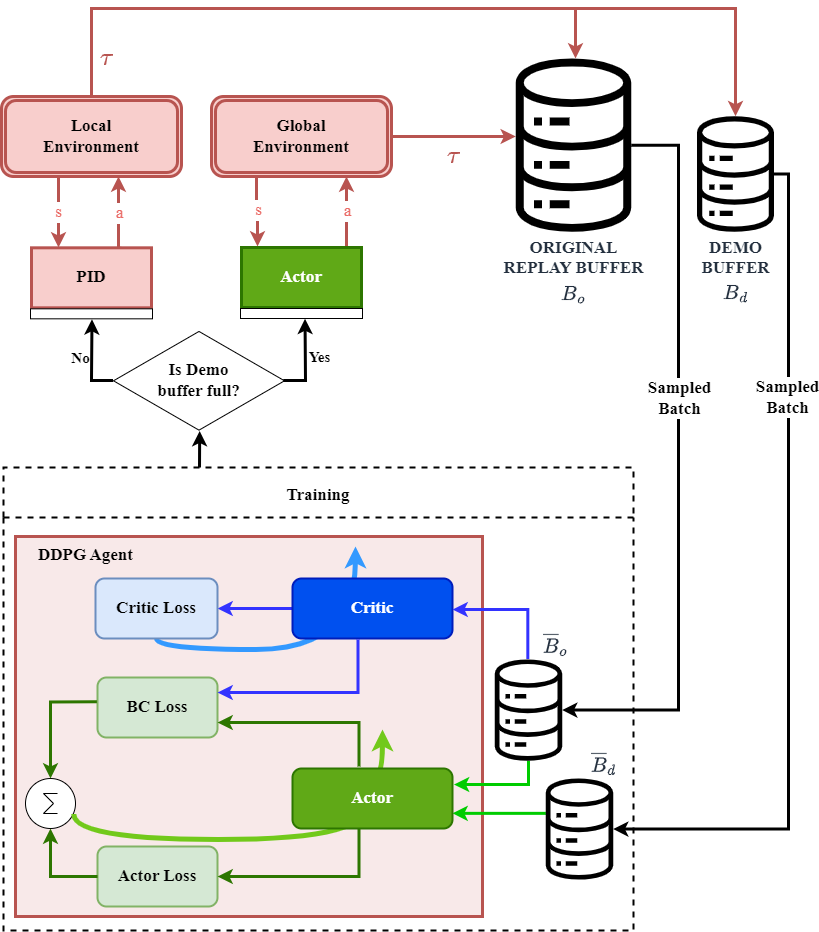}
        \caption{Off-policy based on DDPG}
        \label{fig:arch_ddpg}
    \end{subfigure}
    \caption{Graphical illustration of Demo-EASE (1/2)}
\end{figure}

\begin{figure}[htbp]
\ContinuedFloat
\centering
    \begin{subfigure}[b]{\textwidth}
        \centering
        \includegraphics[width=0.96\textwidth]{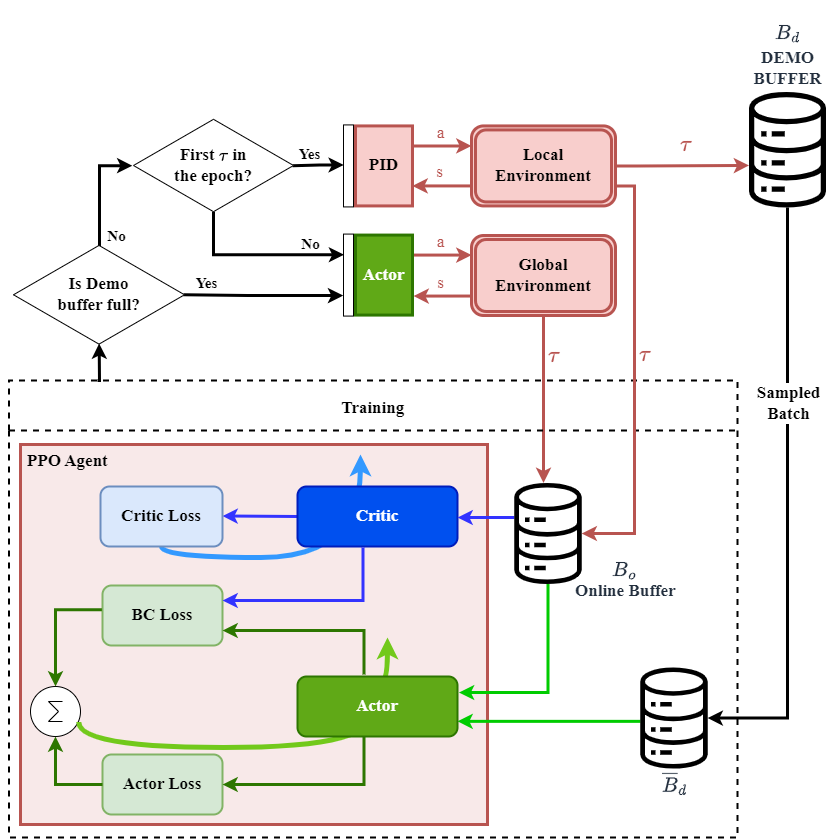}
        \caption{On-policy based on PPO}
        \label{fig:arch_ppo}
    \end{subfigure}
    \caption{Graphical illustration of Demo-EASE (2/2)}
    \label{fig:demo-ease-part2}
\end{figure}

The training algorithms of the Demo-EASE framework for off-policy and on-policy agents are presented in Algorithm \ref{ag:d-ease-off} and Algorithm \ref{ag:d-ease-on}, respectively. In Algorithm \ref{ag:d-ease-off}, $N_{\mathrm{eq}} \in \mathbb{N}^+$ is the number of episodes, $N_{\mathrm{up}} \in \mathbb{N}^+$ is the number of epochs after which the networks are updated, $T \in \mathbb{N}^+$ is the maximum length of each episode, $\sigma \in \mathbb{R}^+$ is the standard deviation of the action exploration, and $\omega \in \mathbb{R}^+$ is the coefficient of target network updates.

\begin{algorithm}[htbp]
\caption{Off-policy Demo-EASE Training based on DDPG}\label{ag:d-ease-off}

\KwSty{Input:}
Parameters $\mu$, $\nu$ for initial policy $\pi_{\mu}$ and action-value function $Q_{\nu}$\;
Parameters $\mu' \!\leftarrow\! \mu$, $\nu' \!\leftarrow\! \nu$ for target networks $\pi'_{\mu'}$ and $Q'_{\nu'}$\;
Empty replay buffer $\mathcal{B}_{\mathrm{o}}$, and full replay buffer $\mathcal{B}_{\mathrm{d}}$\;
\KwSty{Output:} trained policy $\pi_{\mu}$ and action-value function $Q_{\nu}$

\For{$i \leftarrow 1$ \KwTo $N_{\mathrm{ep}}$}{
    
    \For{$j \leftarrow 0$ \KwTo $N_{\mathrm{up}}$}{
    
        $s_t \leftarrow q_0$\,, $t \leftarrow 0$\;
        \While{$t<T$ \& $\zeta_t=0$}{
        Observe $s_t$ and execute $a_t \sim \mathcal{N}\!\left(\pi_\mu(s_t), \sigma \right)$\;
        Observe $s'_t$, $r_t$ and determine $\zeta_t$\;
        Store sample $\left\{s_t, a_t, r_t, s'_t, \zeta_t \right\} \rightarrow \mathcal{B}_{\mathrm{o}}$\;
        $s_t \leftarrow s_t'$\,, $t \leftarrow t+1$\;
        }
    }
    
    Sample batch buffers $\overline{\mathcal{B}}_{\mathrm{o}} \sim \mathcal{B}_{\mathrm{o}}$, $\overline{\mathcal{B}}_{\mathrm{d}} \sim \mathcal{B}_{\mathrm{d}}$\;

    Calculate $\mathcal{L}_O$ using \\
    $\mathcal{L}_{O}\!\left(\overline{\mathcal{B}}_{\mathrm{o}}\right) = -\frac{1}{\overline{N}_{\mathrm{o}}} \sum_{t=0}^{\overline{N}_{\mathrm{o}}-1} Q(s_t, \pi(s_t))$\;
    
    Calculate $\mathcal{L}_{BC}$ using Equation \ref{eq:bc_gain}\;
    
    Calculate $\mathcal{L}_A$ using Equation \ref{eq:loss_actor} and $\nabla_\mu \mathcal{L}_A$\;

    Calculate $\mathcal{L}_C$ and $\nabla_\nu\mathcal{L}_C$ using \\
    $\mathcal{L}_C\!\left(\overline{\mathcal{B}}_{\mathrm{o}}\right) = \frac{1}{\overline{N}_{\mathrm{o}}} \sum_{t=0}^{\overline{N}_{\mathrm{o}}-1} \left(r_t  - Q(s_t, \pi(s_t)) \right. + \left.\gamma(1-\zeta_t)Q'(s'_t, \pi'(s'_t))  \right)^2$\;
    
    Update networks $\pi_{\mu}, Q_{\nu}$ by gradient descent:\\
    $\mu \leftarrow \mu + \alpha_{\mu} \nabla_\mu \mathcal{L}_A$,\
    $\nu \leftarrow \nu + \alpha_{\nu} \nabla_\nu \mathcal{L}_C$\;
    
    Update the target networks $\pi_{\mu'}, Q_{\nu'}$:\\
        $\mu' \leftarrow \omega \mu' + (1 - \omega) \mu ,\ \nu' \leftarrow \omega \nu' + (1 - \omega) \nu$\;
}
\end{algorithm}

\begin{algorithm}[htbp]
\caption{On-policy Demo-EASE Training based on PPO-Clip}\label{ag:d-ease-on}

\KwSty{Input:}
Parameters $\omega$, $\nu$ for initial policy $\pi_{\omega}$ and value $V_{\nu}$ networks\;
Empty replay buffers $\mathcal{B}_{\mathrm{o}}, \mathcal{B}_{\mathrm{d}}$\;
\KwSty{Output:} trained policy $\pi_{\omega}$ and value $V_{\nu}$

\For{$i \leftarrow 1$ \KwTo $N_{\mathrm{ep}}$}{

    \For{$j \leftarrow 0$ \KwTo $N_{\mathrm{o}}$}{
        \eIf{j $\leq N_{\mathrm{o}}$ $\mathbf{and}\, \mathcal{B}_{\mathrm{d}}  \text{ is not } \mathbf{full}$}{
        Run a demonstration episode and store the trajectory $\rightarrow \mathcal{B}_{\mathrm{d}}, \mathcal{B}_{\mathrm{o}}$\;
        }
        {
        $s_t \leftarrow q_0$\,, $t \leftarrow 0$\;
        \While{$t<T$ \& $\zeta_t=0$}{
        Observe $s_t$ and execute $a_t \sim \mathcal{N}\!\left(\mu_{\omega_i}(s_t), \sigma_{\omega_i}(s_t) \right)$\;
        Observe $s_{t+1}$, $r_t$ and determine $\zeta_t$\;
        $s_t \leftarrow s_{t+1}$\,, $t \leftarrow t+1$\;
        }
        Store the trajectory $\rightarrow \mathcal{B}_{\mathrm{o}}$\;
        }
    }

    Compute Rewards-to-go for the trajectories in $\mathcal{B}_{\mathrm{o}}$, $\hat{R}_t = \sum_{l=t}^{T} r_l$;
    
    Sample batch buffers $\overline{\mathcal{B}}_{\mathrm{o}} \sim \mathcal{B}_{\mathrm{o}}$, $\overline{\mathcal{B}}_{\mathrm{d}} \sim \mathcal{B}_{\mathrm{d}}$\;
    $k \leftarrow 0$\;
    
    \While{$k \leq N_{\mathrm{up}}$ \textbf{\text{and}} $D_{KL}(\pi_{\omega_{i, k}}, \pi_{\omega_{i, 0}}) < D_{KL}^{\text{target}}$}{
        
        Compute Advantage estimates for the trajectories in $\mathcal{B}_{\mathrm{o}}$ using the Generalized Advantage Estimator (GAE): \\
        $\hat{A}^{GAE(\gamma, \lambda)}_t = \sum_{l=t}^{T-1} (\gamma \lambda)^{l-t} (r_l + \gamma V_{\omega_i, k}(s_{l+1}) - V_{\omega_i, k}(s_l)) $\\
        where $\gamma$ is the discount factor for the MDP and $\lambda$ is the advantage estimation discount factor;

        Calculate \\
        $\mathcal{L}_{O} = -\frac{1}{N_{\mathrm{o}}T} \sum_{\mathcal{B}_\mathrm{o}}\sum_{t=0}^{T} \min\left(r^\omega_t \hat{A}_t, \mathrm{clip}(r^\omega_t, 1-\epsilon, 1+\epsilon)\hat{A}_t\right)$ \\
        with $r^\omega_t = {\pi_\omega(a_t|s_t)}/{\pi_{\omega_{i, k}}(a_t|s_t)}$ and $\epsilon$ as the clipping threshold\;
        
        Calculate $\mathcal{L}_{BC}$ using  Equation \ref{eq:bc_gain}\;
        
        Calculate $\mathcal{L}_A$ using Equation \ref{eq:loss_actor} and $\nabla_\omega \mathcal{L}_A$\;
        
        Calculate $\mathcal{L}_C$ and $\nabla_\nu\mathcal{L}_C$ using \\
        $\mathcal{L}_C = \frac{1}{N_{\mathrm{o}}T} \sum_{\mathcal{B}_\mathrm{o}}\sum_{t=0}^{T} \left(V_\nu(s_t) - \hat{R}_t\right)^2$\;
    
        Update networks $\pi_{\omega}, V_{\nu}$ by gradient descent:\\
        $\omega_{i, k+1} \leftarrow \omega_{i, k} + \alpha_{\omega} \nabla_\omega \mathcal{L}_A$,\
        $\nu_{i, k+1} \leftarrow \nu_{i, k} + \alpha_{\nu} \nabla_\nu \mathcal{L}_C$\;

        $k \leftarrow k+1$;        
    }
    
}
\end{algorithm}

\subsection{Masked Behavior Cloning}\label{sec:loss}

This section introduces the loss functions designed to train the RL agents. 
We do not use all the samples in the replay buffers $\mathcal{B}_{\mathrm{o}}$ and $\mathcal{B}_{\mathrm{d}}$ to update the actor and critic networks of our RL agents. Instead, we use two smaller batches $\overline{\mathcal{B}}_{\mathrm{o}}$ and $\overline{\mathcal{B}}_{\mathrm{d}}$ which contain the shuffled and randomly sampled transitions of $\mathcal{B}_{\mathrm{o}}$ and $\mathcal{B}_{\mathrm{d}}$, respectively. The sizes of $\overline{\mathcal{B}}_{\mathrm{o}}$ and $\overline{\mathcal{B}}_{\mathrm{d}}$ are respectively $\overline{N}_{\mathrm{o}}$ and $\overline{N}_{\mathrm{d}}$ which satisfy $\overline{N}_{\mathrm{d}} < N_{\mathrm{d}}$ and $\overline{N}_{\mathrm{o}} < N_{\mathrm{o}}$. The conventional critic loss function is minimized to update the critic networks in both off-policy and on-policy versions of Demo-EASE.
The actor networks, however, are updated using both batches $\overline{\mathcal{B}}_{\mathrm{o}}$ and $\overline{\mathcal{B}}_{\mathrm{d}}$ with a combined loss function
\begin{equation}\label{eq:loss_actor}
\textstyle \mathcal{L}_{A}\!\left(\overline{\mathcal{B}}_{\mathrm{o}}, \overline{\mathcal{B}}_{\mathrm{d}}\right) = \mathcal{L}_{O}\! \left( \overline{\mathcal{B}}_{\mathrm{o}} \right) + \lambda_{BC} \mathcal{L}_{BC}\!\left( \overline{\mathcal{B}}_{\mathrm{d}} \right),
\end{equation}
where $\mathcal{L}_O$ is the original actor loss function based on either DDPG or PPO. $\mathcal{L}_{BC}$ is the behavior cloning loss function used to encourage the compatibility between the policies, and $\lambda_{BC}$ is a hyperparameter to adjust the regulation. The behavior cloning loss function $\mathcal{L}_{BC}$, inspired by the behavior cloning in imitation learning~\cite{nair2018overcoming}, is defined as 
\begin{equation}\label{eq:bc_gain}
    \textstyle \mathcal{L}_{BC}\!\left(\overline{\mathcal{B}}_{\mathrm{d}}\right) = \frac{1}{\overline{N}_{\mathrm{d}}}\sum_{t=0}^{\overline{N}_{\mathrm{d}}-1}M_t\left(a_t^{\mathrm{d}}-\hat{a}_t \right)^2,
\end{equation}
where $\hat{a}_t= \pi\!\left(s_t^{\mathrm{d}} \right)$ is the agent action subject to the current policy $\pi$ and the demonstration state $s^{\mathrm{d}}_t$, $\left(s_t^{\mathrm{d}}, a_t^{\mathrm{d}}\right) \sim \overline{\mathcal{B}}_{\mathrm{d}}$. Note that in the on-policy method, we are using a variational actor that produces a mean $\mu$ and variance $\sigma$ for actions, so $\hat{a}_t= \mu\!\left(s_t^{\mathrm{d}} \right)$. $M_t$ is a binary mask that indicates whether the demonstration action $a^{\mathrm{d}}_t$ is superior to the current policy action $a_t$. The superiority in the case of the off-policy version is defined as a higher state-action pair value for $a^{\mathrm{d}}_t$, i.e.,
\begin{equation}\label{eq:mask_q}
    Q(s_t^{\mathrm{d}}, \hat{a}_t) < Q(s_t^{\mathrm{d}}, a^{\mathrm{d}}_t).
\end{equation}
Alternatively, in the on-policy version, the advantage of behavior cloning is defined as when the immediate reward received by the demonstration policy for $a^{\mathrm{d}}_t$ is higher than the estimated state value
\begin{equation}\label{eq:mask_v}
    V(s_t^{\mathrm{d}}) < R(s_t^{\mathrm{d}}, a^{\mathrm{d}}_t).    
\end{equation}
The purpose of $\mathcal{L}_{BC}$ is to balance between the demonstration and the original experience of the agent. In the initial stages of the training, when the policy is not yet trained, $\pi$ imitates the demonstration policy $\pi^{\mathrm{d}}$. When the agent is well-trained such that its actor performs better than the demonstration, $\mathcal{L}_{BC}$ vanishes and the demonstration policy $\pi^{\mathrm{d}}$ is ignored. The behavior cloning gain hyperparameter, $\lambda_{BC}$, adjusts the extent of cloning. In other words, $\lambda_{BC}$ balances between the training speed and the generalizability of the agent.

\section{Experiment Setup}\label{sec:exp}
\begin{figure}[htbp]
    \centering
    \includegraphics[width=0.7\linewidth,clip]{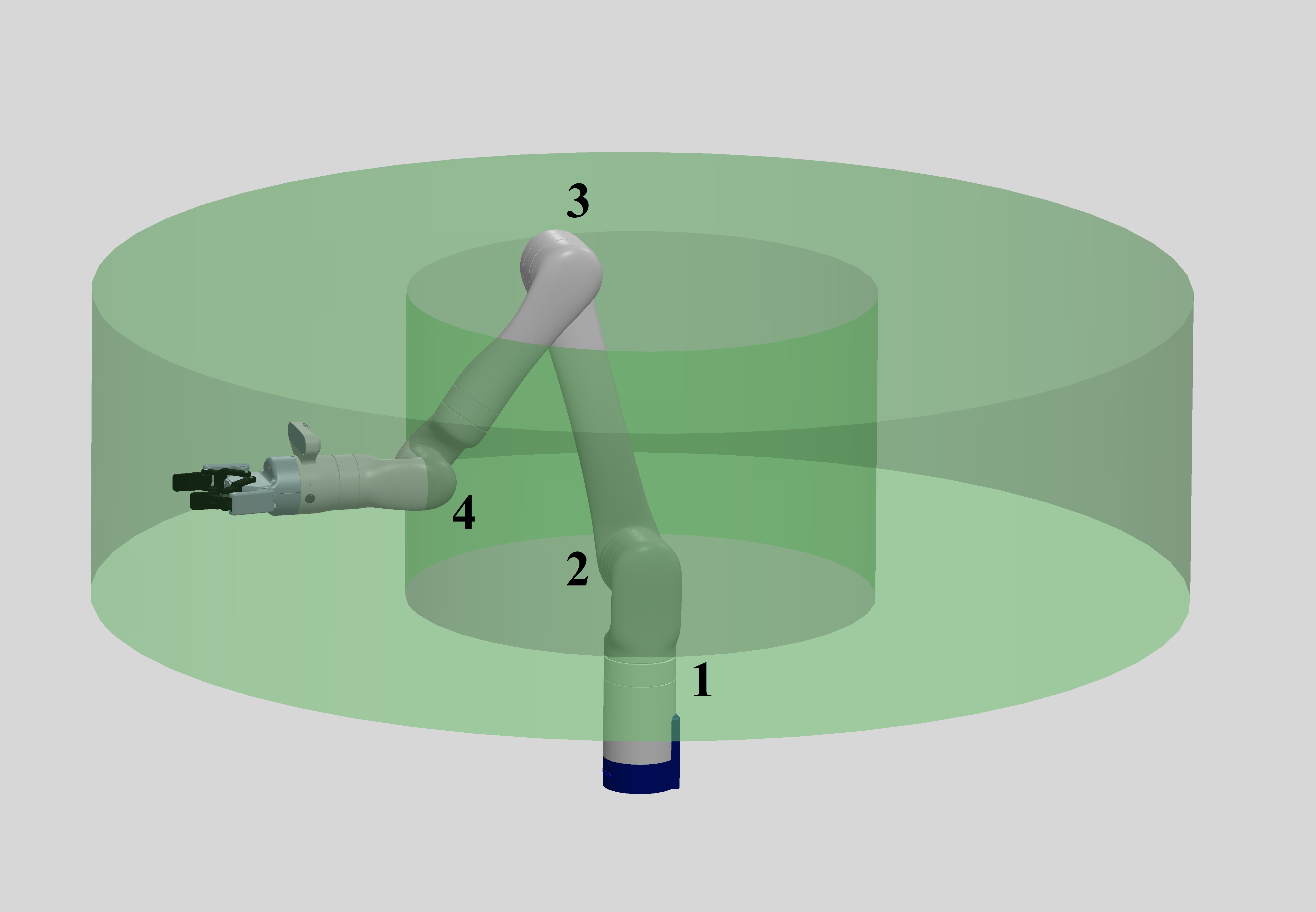}
    \caption{The 6-DOF Kinova\textregistered~Gen3 robot model. The workspace is the green tube showing $\min$ and $\max$ extremities of polar radius and height. The four active joints are labeled as well.}
    \label{fig:workspace}
\end{figure}

To validate the proposed method, we set up a simulation experiment in the PyBullet environment using a 6-DOF Kinova\textregistered~Gen3 robot model with a Robotiq\textregistered~2F-85 gripper as shown in Figure \ref{fig:workspace}. Since our experiments do not involve complex dexterity, only four actuators (\#\,1, \#\,2, \#\,3, and \#\,4) are active and the other two are locked, as shown in Figure \ref{fig:workspace}, which results in a 4-DOF active robot arm. The experimental simulation is performed on a computer workstation equipped with an AMD Ryzen 9 5900X CPU, an Nvidia RTX 3090 GPU, and 64GB (2x32) Corsair DDR4 memory, with Ubuntu 18.04 LTS operating system. We use \textit{spinningup}~\cite{SpinningUp2018} as the DDPG and PPO benchmarks to develop our method. Our implementation is available here: \url{https://www.github.com/amsoufi/DemoEASE}.

The configurations of the three tasks are interpreted as follows. Note that we will use cylinder coordinates $(\theta, \rho, z)$ to represent the position in the workspace for the remaining part of this paper.

\subsection{Environments}
\begin{figure}[ht]
    \centering
    \begin{subfigure}[b]{0.64\textwidth}
        \includegraphics[width=\linewidth]{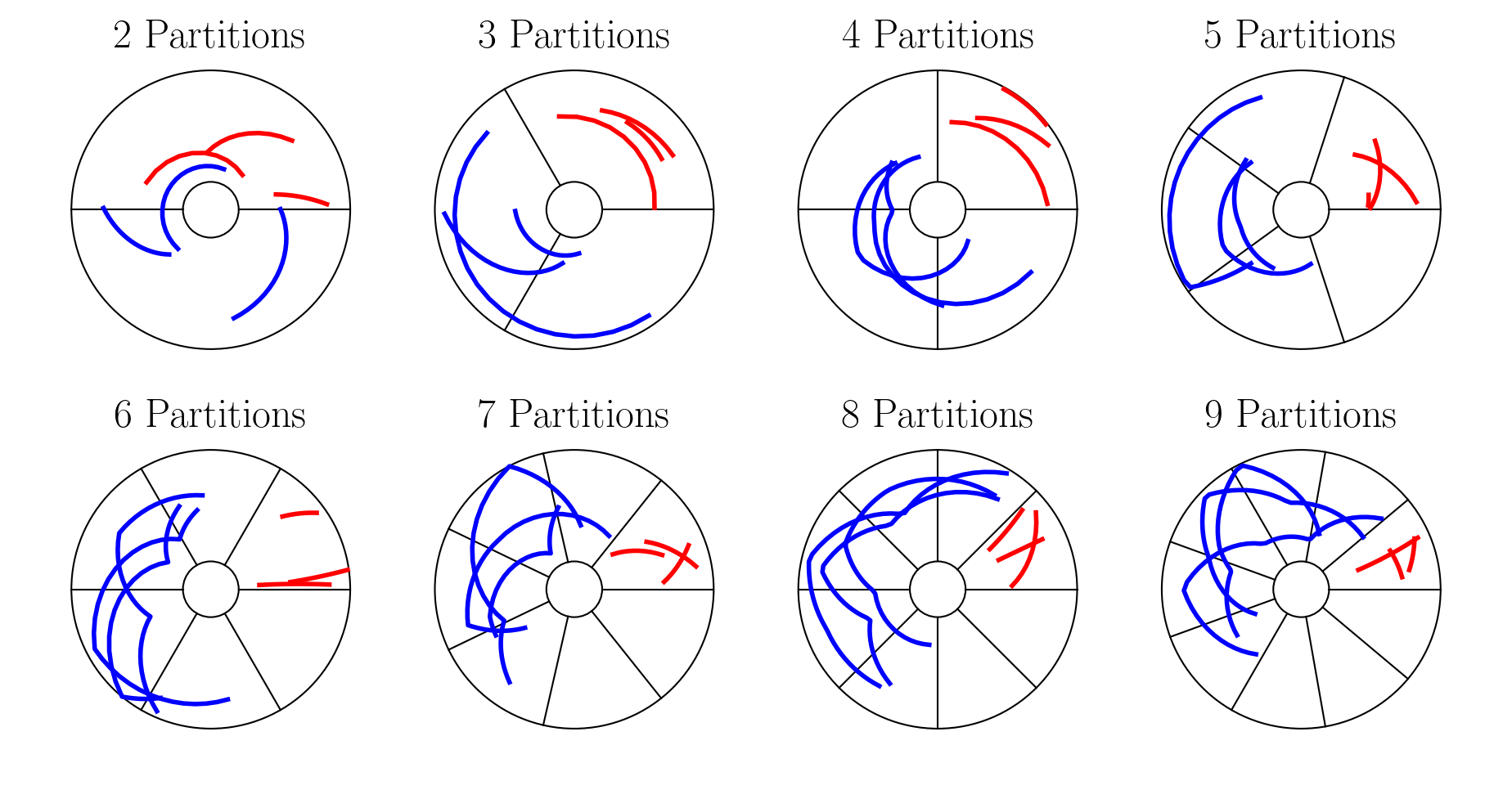}
    \end{subfigure}%
    \hfill
    \begin{subfigure}[b]{0.35\textwidth}
        \centering
        \begin{tikzpicture}
            \node[anchor=south west, inner sep=0] (image) at (0,0) {\includegraphics[width=\linewidth]{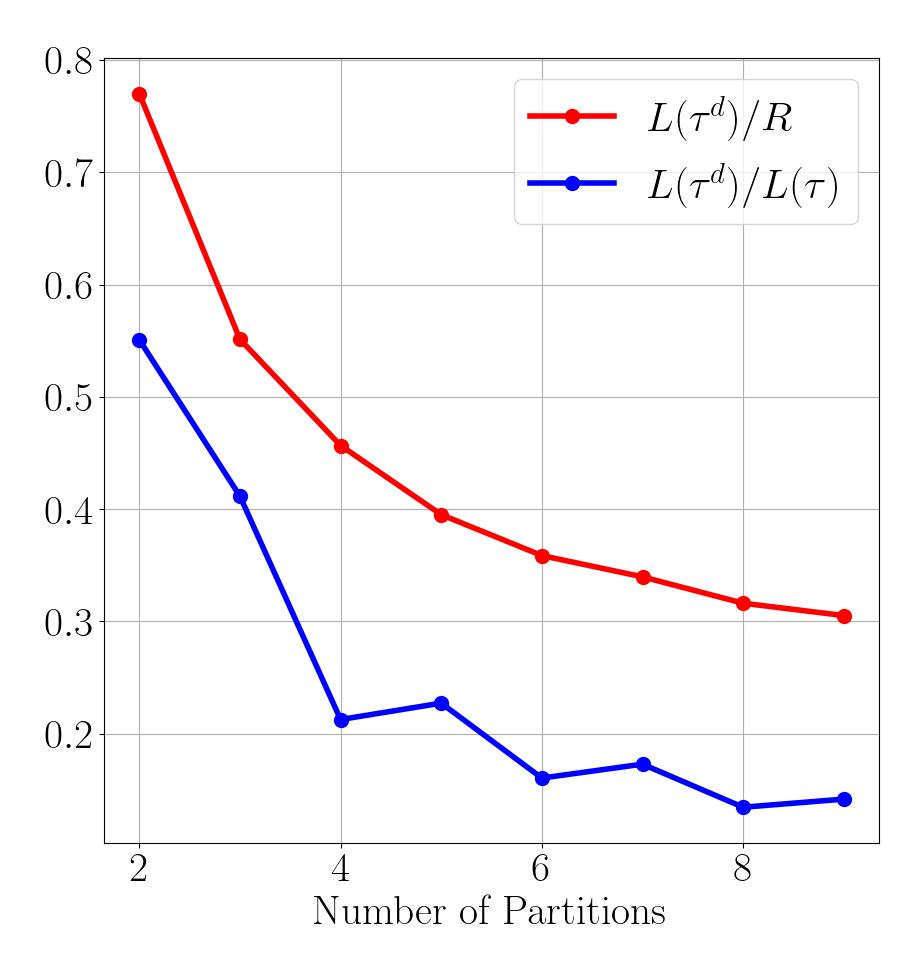}};
            \begin{scope}[x={(image.south east)}, y={(image.north west)}]
                \draw[purple, thick, rounded corners] (0.32, 0.22) rectangle (0.42, 0.57);
            \end{scope}
        \end{tikzpicture}
    \end{subfigure}
    \caption{The effect of different partition numbers in experiments. \textbf{Left:} The schematic illustration of paths in the local and global environment. \textbf{Right:} The normalized average demonstration length $L(\tau^d)/R$ and its ratio to the average regular path length $L(\tau^d)/L(\tau)$. $R$ is the outer radius of the workspace. Our choice with $K=4$ partitions is indicated.}
    \label{fig:partitions}
\end{figure}

The number of partitions $K$ used to define the symmetrically partitionable space is a design choice in the Demo-EASE framework. In our environments, we fix $K=4$ to maintain a balance between abstraction and coverage. A larger $K$ would result in excessive partitioning. This limits the initial knowledge transfer and delays bootstrapping of the RL agent. In other words, demonstrating only \textit{near-goal} states cannot effectively propagate to the bigger state space of the global environment. Conversely, using too few partitions requires a more proficient demonstration generation. Hence, the performance of the demonstration policy might degrade in larger partitions, as shown by the reported success rate of the demonstration policy in the following sections. On the other hand, using better controllers to overcome this issue would dilute the significance of leveraging symmetry, shifting the paradigm toward imitation learning. For these reasons, although the role of $K$ is not empirically studied here, the chosen value of $K=4$ reflects a practical. Figure \ref{fig:partitions} is an abstract schematic of the influence of partition number on path lengths. The paths are in a 2-D environment as an abstract top-view of the robot workspace (similar to Figure \ref{fig:workspace}). The long paths start from one partition and end in the furthest opposite partition, such that they cover the whole workspace. Intermediate waypoints are picked randomly in each partition on the way. Trajectory is then linearly interpolated, in a cylindrical coordinate system, between each two consecutive waypoints. This is inspired by the path planning strategy for creating demonstrations with the PID controller outlined in Equation \ref{eq:pid}.

In a symmetrically partitionable space, the demonstration policy is executed uniformly in all local environments. Apart from a small noise, we fix the initial position $q_0$ for each local environment, and sample the goal and object positions $P_g$ and $P_o$ randomly from uniform distributions.
For each episode of the chosen tasks, the robot moves from an initial joint angle $q_0 \in \mathbb{R}^n$ to reach the goal position $P_g \in \mathbb{R}^3$. For different stages, demonstration recording, training, and testing, the initial joint angle $q_0$ and the goal position $P_g$ are sampled differently to evaluate the generalizability of the agent. The P2P-O and P\&P tasks also involve the sampling of the object position $P_{o}$, which is sampled according to the initial robot configuration. The demonstration is recorded in local environments $\mathcal{S}_k$, where $k \sim \left\{ 1,2,3,4 \right\}$ is uniformly sampled, while the training and test are performed in the global environment $\mathcal{S}$.

Although the Demo-EASE framework is developed based on a formal notion of symmetry (GAS), the environments used in our experiments show possible extension to environments that are not strictly globally symmetric. In particular, the P2P-O task includes an obstacle that is also replicated in every subregion, allowing us to treat each subdomain similarly through abstraction. The workspace is abstracted into four partitions in our design, where the same short demonstration can be virtually replayed with symmetric transformations. The demonstrations can therefore be generated in one asymmetric subregion and then projected into others using the mappings introduced for GAS. This approach allows for introducing symmetry-breaking elements while benefiting from symmetry-guided abstraction. The agent receives behavior cloning guidance on paths, encompassing all asymmetric elements in one specific partition, and explores to generalize to other parts of the space.

\subsubsection{Configuration of the P2P task}
For the P2P task, we chose the Off-policy Demo-EASE to train the RL agent. The initial position of the end effector is sampled uniformly such that $\theta_0 \sim \left\{\frac{k\pi}{2} - \frac{\pi}{4} \right\}$,~$\rho_0 = 0.52$,~$z_0 = 0.42$,
for all $k \sim \left\{ 0,1,2,3 \right\}$. Here, we use the polar coordinate $(\theta_0, \rho_0, z_0)$ to represent the end-effector position in the workspace.
Then, inverse kinematics is calculated to obtain the initial joint angles $q_0$. The goal position $(\theta_g,\rho_g,z_g)$ is uniformly sampled from $\theta_g \sim \left[\frac{k\pi}{2} - \frac{\pi}{2}, \frac{k\pi}{2} \right)$, $\rho_g \sim [\,0.4,\, 0.6\,)$, $
z_g \sim [\,0.35, \, 0.55\,)$,
which is then transformed to the Cartesian coordinate $P_g$. In such a manner, the sampled goal is restricted to the same local environment as the initial position of the robot end-effector. The number of the demonstration episodes $N_{\mathrm{demos}}$ is selected as 0, 80, and 160 for the comparison study. Four examples of P2P demonstrations are illustrated in Figure \ref{fig:p2p_demo}.

\subsubsection{Configuration of the P2P-O and P\&P}
\begin{figure}[htbp]
\centering
    \begin{subfigure}[b]{0.48\textwidth}
        \centering
        \includegraphics[width=0.95\textwidth,clip]{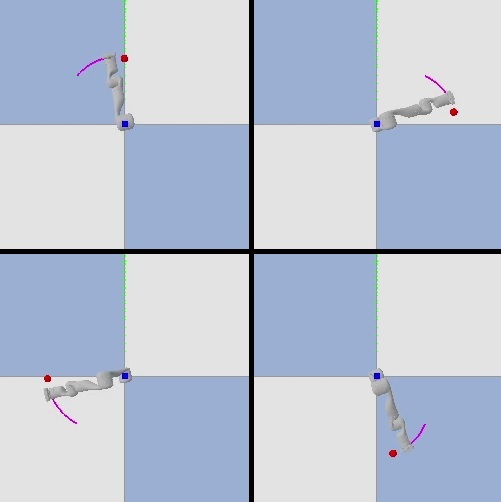}
        \caption{P2P}
        \label{fig:p2p_demo}
    \end{subfigure}
    \begin{subfigure}[b]{0.48\textwidth}
        \centering
        \includegraphics[width=0.95\linewidth,clip]{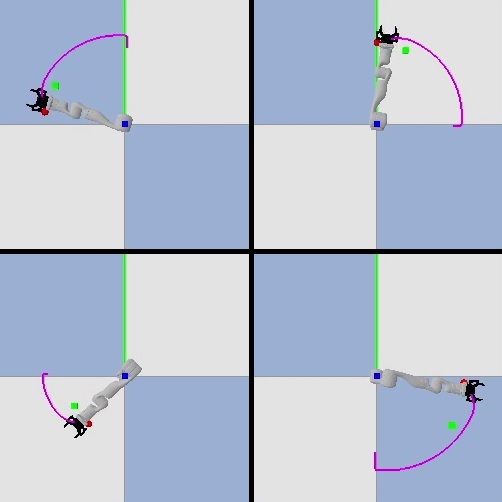}
        \caption{P2P with obstacle}
        \label{fig:p2p_obs_demo}
    \end{subfigure}
    \caption{\small{Samples of the demonstrations used to prepare the local demo replay buffers}}
    \label{fig:demo}
\end{figure}

The P2P-O task was trained using Off-policy Demo-EASE while we chose the On-policy version for P\&P. For both tasks, the initial position of the robot end-effector is uniformly sampled from $\theta_0 \sim \left\{ \frac{k\pi}{2} \right\}$,~$\rho_0 = 0.52$,~$z_0 = 0.42$, for all $k \sim  \{0, 1, 2, 3\}$, which is then transformed to the initial joint angle $q_0$. The target position $P_g$ is sampled from
$\theta_g \sim \left[\frac{k\pi}{2} - \frac{\pi}{4}, \frac{k\pi}{2}\right)$,
$\rho_g \sim [\,0.4, \,0.6\,)$,
$z_g \sim [\,0.35, \, 0.55\,)$.
A cube sized $0.04\,$m is placed at position $P_o$ which is determined based on the goal position so that it is located between the initial position of the gripper and the target,
$\theta_o \sim \theta_g + \left\{ -1,1 \right\} \times \left[\frac{\pi}{12},\frac{\pi}{6}\right)$,
$z_o \sim [\,z_g - 0.1, z_g + 0.1\,)$.
The subscript $o$ indicates the coordinate of the object. The number of demonstration episodes $N_{\mathrm{demos}}$ is set at 100, 200, and 400 for P2P-O and 150, 250, and 400 for P\&P. They are larger than what is chosen in the P2P task due to their complexity. Four examples of P2P-O demonstrations are shown in Figure \ref{fig:p2p_obs_demo}.

\subsection{Agent Training}\label{sec:def_reward}
The training of the agent is performed in the global environment $\mathcal{S}$. For both the P2P and the P2P-O tasks, the initial position of the robot end-effector is fixed in 
$\theta_0 = \pi/4$,~$\rho_0 = 0.52$,~$z_0 = 0.42$,
and the target positions $P_{g}$ is sampled from
\begin{equation}\label{eq:goal-p2p}
\theta_g \sim [\,-\pi,\, \pi\,) ,\
\rho_g \sim [\,0.3, \, 0.7\,) ,\
z_g \sim [\,0.25, \, 0.65\,).
\end{equation}
For the P2P-O and P\&P tasks, the object position is sampled from
$\theta_b = \theta_g + \mathrm{sgn}(\theta_1 - 0.5)\theta_2$, $\rho_b \sim [\,0.4, \,0.6\,)$, $z_b \sim [z_g - 0.1, z_g + 0.1)$,
where $\mathrm{sgn}(\cdot)$ is the sign function that produces $1$, $-1$, and $0$ for positive, negative, and zero values, respectively. $\theta_1 \sim [\,0,~1\,)$ and $\theta_2 \sim \left[\,{\pi}/{12}, {\pi}/{6} \, \right)$ are random variables sampled from uniform distributions. Such a definition of $\theta_b$ is to ensure that the block is forced to be towards a clockwise or counterclockwise direction from the target in the global space.

The reward functions $R$ are selected according to the following concerns.
\begin{enumerate}
\item \textbf{Approaching:} Penalize the reaching error at each time.
\item \textbf{Effort:} Encourage the agent to apply minimum effort.
\item \textbf{Reaching:} Encourage ultimately reaching the goal.
\item \textbf{Collision:} Penalize any collisions with the object.
\item \textbf{Grasping:} Encourage the agent to pick up the object when intended.
\end{enumerate}

Thus, we design the following reward function,
\begin{equation}
    R(s_t,a_t) := r^{\mathrm{dst}}_t + r^{\mathrm{eft}}_t + r^{\mathrm{rch}}_t + r^{\mathrm{cls}}_t + r^{\mathrm{grsp}}_t.
\end{equation}
The first term $r^{\mathrm{dst}}_t = -\alpha_1 || e_t ||$ is to penalize the reaching error $e_t$, where $\alpha_1 \in \mathbb{R}^+$ is design parameter to be determined. This term stimulates the reaching towards the goal. The second term $r^{\mathrm{eft}}_t = -\alpha_2 \left\| \hat{\tau}_t \right\|$ is to penalize the actuation torques of the robot, 
where $\alpha_2$ is a parameter and $\hat{\tau}_t = \left[\hat{\tau}_t^{(1)} ~ \hat{\tau}_t^{(2)} ~ \hat{\tau}_t^{(3)} ~ \hat{\tau}_t^{(4)} \,\right]^{\top}$ are the normalized actuation torques of the four active joints of the robot. For each joint $j$, $\hat{\tau}_t^{(j)} = \lvert \tau_t^{(j)} \rvert / \tau^{(j)}_{\max}$, where $\tau_t^{(j)}$ is the actual actuation torque at time $t$ and $\tau^{(j)}_{\max}$ is the maximum torque, where $\tau^{(1,2,3)}_{\max} = 39\,$N$\cdot$m and $\tau^{(4)}_{\max} = 9\,$N$\cdot$m.
The third term $\displaystyle r^{\mathrm{rch}}_t = 
    \left\{\begin{array}{ll}
    0, & \text{if}\, \left\| e_{t} \right\| \geq \varepsilon\\ 
    R_1, & \text{else}
\end{array}\right.$ gives a big positive reward for ultimately reaching the goal, 
where $\varepsilon = 0.05\,$m is the tolerated reaching error. The term $\displaystyle r^{\mathrm{cls}}_t = 
    \left\{\begin{array}{ll}
    -R_2, & \text{if collision detected} \\ 
    0, & \text{else}
\end{array}\right.$ exerts a big negative reward for collision either with the object or the robot itself, 
and the last term $\displaystyle r^{\mathrm{grsp}}_t = 
    \left\{\begin{array}{ll}
    R_3, & \text{if}\, \left\| P_t^e - P_t^o \right\| \leq \varepsilon ~ \text{and} ~ d(\textbf{q}_t^e - \textbf{q}_t^o) \leq \varepsilon_q \\ 
    0, & \text{else}
\end{array}\right.$ is a one-time reward in the P\&P task when the robot reaches the object in a proper position with proper orientation, where $\varepsilon, R_1, R_2, R_3 \in \mathbb{R}^+$ are parameters. For the reward function, the parameters are selected as $\alpha_1 = 2\times10^{-3}$, $\alpha_2 = 10^{-3}$, $R_1=10$, $R_2=R_3=2$ wherever needed for each task. In the definition of $r^{\mathrm{grsp}}_t$, $d(\textbf{q}_t^e - \textbf{q}_t^o)$ is the distance between the end-effector and object quaternions. Also, if the robot satisfies the condition for receiving $R_3$, the grasping occurs automatically until the end of the episode.

The architecture of the actor and critic networks in P2P and P2P-O are multilayer perceptron networks with $[\,128,~512,~128\,]$ and $[\,256,~1024,~256\,]$ units, respectively. A similar architecture with $[\,512,~512,~256\,]$ units is used for both actor and critic networks in P\&P. The remaining parameters of the agent training are indicated in Tables \ref{tab:reward} and \ref{tab:reward2}.

\renewcommand{\arraystretch}{1.15}
\begin{table}[htbp]
\caption{Training parameters for off-policy Demo-EASE experiments}
\centering
\begin{tabular}{clcc}
\toprule
Notation & Parameter & P2P & P2P-O \\
\midrule
$T$ &  Max Episode Length & \multicolumn{1}{c|}{400} & 500 \\
$N_{\mathrm{ep}}$ & Number of epochs & \multicolumn{1}{c|}{250} & 500 \\
$N_{\mathrm{up}}$ & Update interval & \multicolumn{1}{c|}{20} & 25 \\
$N_{\mathrm{o}}$  & Size of $\mathcal{B}_{\mathrm{o}}$ & \multicolumn{2}{c}{$10^6$} \\
$\overline{N}_{\mathrm{o}}$ & Size of $\overline{\mathcal{B}}_{\mathrm{o}}$ & \multicolumn{2}{c}{100} \\
$\overline{N}_{\mathrm{d}}$ & Size of $\overline{\mathcal{B}}_{\mathrm{d}}$ & \multicolumn{2}{c}{100} \\
$\gamma$ & Discount factor & \multicolumn{2}{c}{$0.99$} \\
$\sigma$ & Exploration Noise Std. Dev. & \multicolumn{2}{c}{$0.1$} \\
$\omega$ & Target update coefficient & \multicolumn{2}{c}{$0.995$} \\
$\alpha_\mu$ & Actor's learning rate & \multicolumn{2}{c}{$10^{-3}$} \\
$\alpha_\nu$ & Critic's learning rate & \multicolumn{2}{c}{$10^{-3}$} \\
\bottomrule
\end{tabular}
\label{tab:reward}
\end{table}

\renewcommand{\arraystretch}{1.15}
\begin{table}[htbp]
\caption{Training parameters for on-policy Demo-EASE experiment}
\centering
\begin{tabular}{clcc}
\toprule
Notation & Parameter & P\&P \\
\midrule
$T$ &  Max Episode Length & 800 \\
$N_{\mathrm{ep}}$ & Number of epochs & 50 \\
$N_{\mathrm{up}}$ & Policy updates per epoch & 20 \\
$N_{\mathrm{o}}$  & Size of $\mathcal{B}_{\mathrm{o}}$ & 4000 \\
$\overline{N}_{\mathrm{o}}$ & Size of $\overline{\mathcal{B}}_{\mathrm{o}}$ & 800 \\
$\overline{N}_{\mathrm{d}}$ & Size of $\overline{\mathcal{B}}_{\mathrm{d}}$ & 100 \\
$\gamma$ & Discount factor & $0.99$ \\
$\gamma^{GAE}$ & GAE factor & $0.97$ \\
$\epsilon$ & Clip ratio & $0.2$ \\
$D_{KL}^{\text{target}}$ & KL-Divergence threshold & 0.02 \\
\bottomrule
\end{tabular}
\label{tab:reward2}
\end{table}

\section{Results and Analysis}\label{sec:results}
In this section, we analyze the results of our implementations and evaluate the performance of the proposed method. Additional information about the environment configuration and platforms is presented in Section \ref{sec:exp}.

\subsection{Training Results}
To evaluate the influence of different configurations of the method on the performance, we focus on two important hyperparameters, namely the number of demonstrations $N_{\mathrm{d}}$, and the behavior cloning gain $\lambda_{BC}$. For each combination of $N_{\mathrm{demos}}$ and $\lambda_{BC}$, multiple repeats are performed to reduce the effect of randomization. A conventional DDPG and PPO agent is also trained for comparison, corresponding to $N_{\mathrm{demos}}=0 $ and $\lambda=0$ for the corresponding experiments.

Along with the learning curve that shows the changes in accumulated reward over time, we also use the following numerical metrics to evaluate the performance:

\begin{itemize}
\item[-] \textbf{Initial return} $\overline{R}_{10}$: the average return over the initial 10\% of episode steps;
\item[-] \textbf{Ultimate return} $\overline{R}_{90}$: the average return over the final 10\% of episode steps;
\item[-] \textbf{Reward Increment} $I_R = \overline{R}_{90} - \overline{R}_{10}$: the increase of the returns during training;
\item[-] \textbf{Half-trained time} $T_{50}$: the episode at which the return first exceed 50\% of $\overline{R}_{90}$.
\end{itemize}

\subsubsection{The P\&P Task}
\begin{figure}[htbp]
\centering
    \begin{subfigure}[b]{0.7\textwidth}
        \centering
        \includegraphics[width=0.98\textwidth,clip]{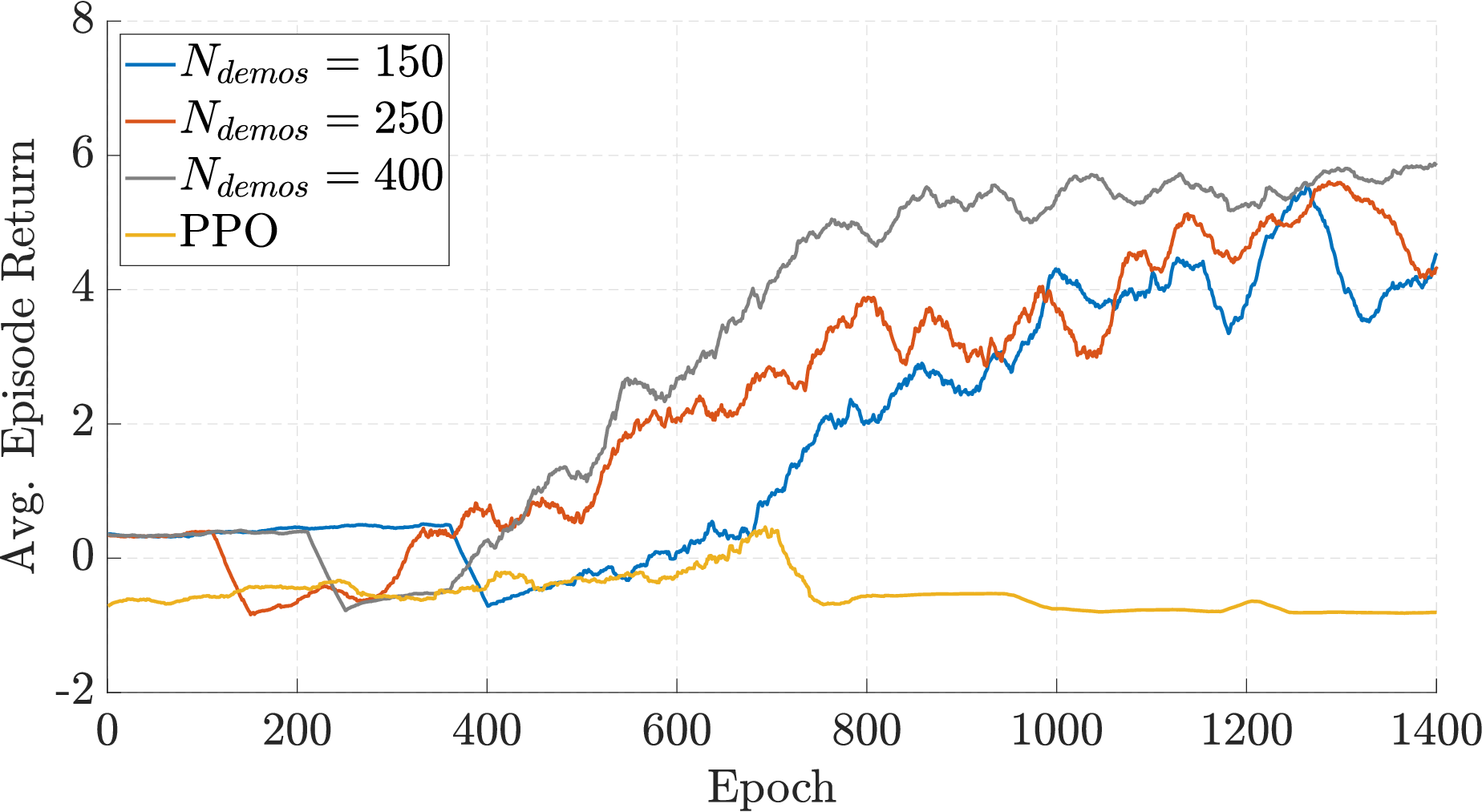}
        \caption{\small{Different $N_{\mathrm{demos}}$ with $\lambda_{BC}=15$}}
        \label{fig:pnp_numdemo}
    \end{subfigure}
    \begin{subfigure}[b]{0.7\textwidth}
        \centering
        \includegraphics[width=0.98\textwidth,clip]{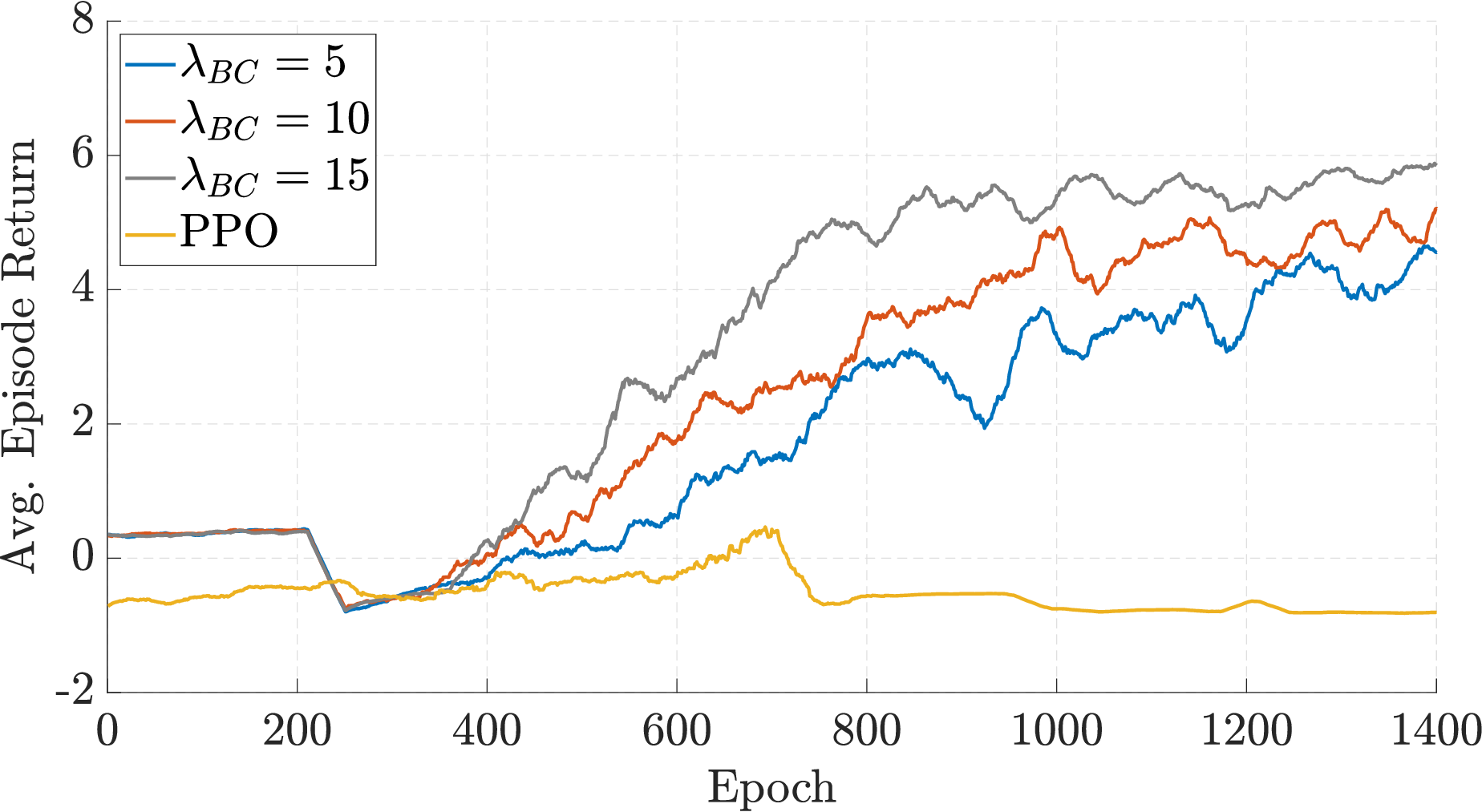}
        \caption{\small{Different $\lambda_{BC}$ with $N_{\mathrm{demos}}=400$}}
        \label{fig:pnp_lambda}
    \end{subfigure}
    \caption{\small{The effect of hyperparameters on the P\&P learning curves}}
    \label{fig:pnp_hyp}
\end{figure}

Figure \ref{fig:pnp_hyp} shows the training curves for the P\&P task. It can be noticed that agents trained using Demo-EASE show a significant advantage over the original baseline. As expected, the learning curves of all Demo-EASE agents, even those with minimal adjustments, start to grow early on. Also, more demonstrations or larger $\lambda_{BC}$ in training leads to higher accumulated rewards during the training process. Additional training metrics for the P\&P task are listed in Table \ref{tab:pnp}.

\begin{table}[htbp]
\begin{center}
\begin{minipage}{0.8\linewidth}
\caption{Numerical training metrics of the P\&P agent}\label{tab:pnp}
\begin{tabular*}{0.96\linewidth}{@{\extracolsep{\fill}}lccccccc@{\extracolsep{\fill}}}
\toprule%
\multirow{2}{*}{Agent} & \multicolumn{3}{@{}c@{}}{$\lambda_{BC}$ ($N_{\mathrm{demos}} = 400$)} & \multicolumn{3}{@{}c@{}}{$N_{\mathrm{demos}}$ ($\lambda_{BC}=15$)} & \multirow{2}{*}{PPO} \\
\cmidrule{2-4}\cmidrule{5-7}%
& 5 & 10 & 15 & 150 & 250 & 400 & \\
\midrule
$\overline{R}_{10}$  &0.36    &0.37    &0.36  &0.35    &0.19   &0.36 &-0.60\\
$\overline{R}_{90}$ &1.67    &2.29    &3.03   &1.73    &2.26    &3.03 &-0.51\\
$I_R$ &1.30    &1.92    &2.67   &1.38    &2.07    &2.67  &0.09\\
$T_{50}$ &649   &612   &548  &721   &572   &548 &655\\
\bottomrule
\end{tabular*}
\end{minipage}
\end{center}
\end{table}

\subsubsection{The P2P Task}
\begin{figure}[htbp]
\centering
    \begin{subfigure}[b]{0.7\textwidth}
        \includegraphics[width=0.98\textwidth,clip]{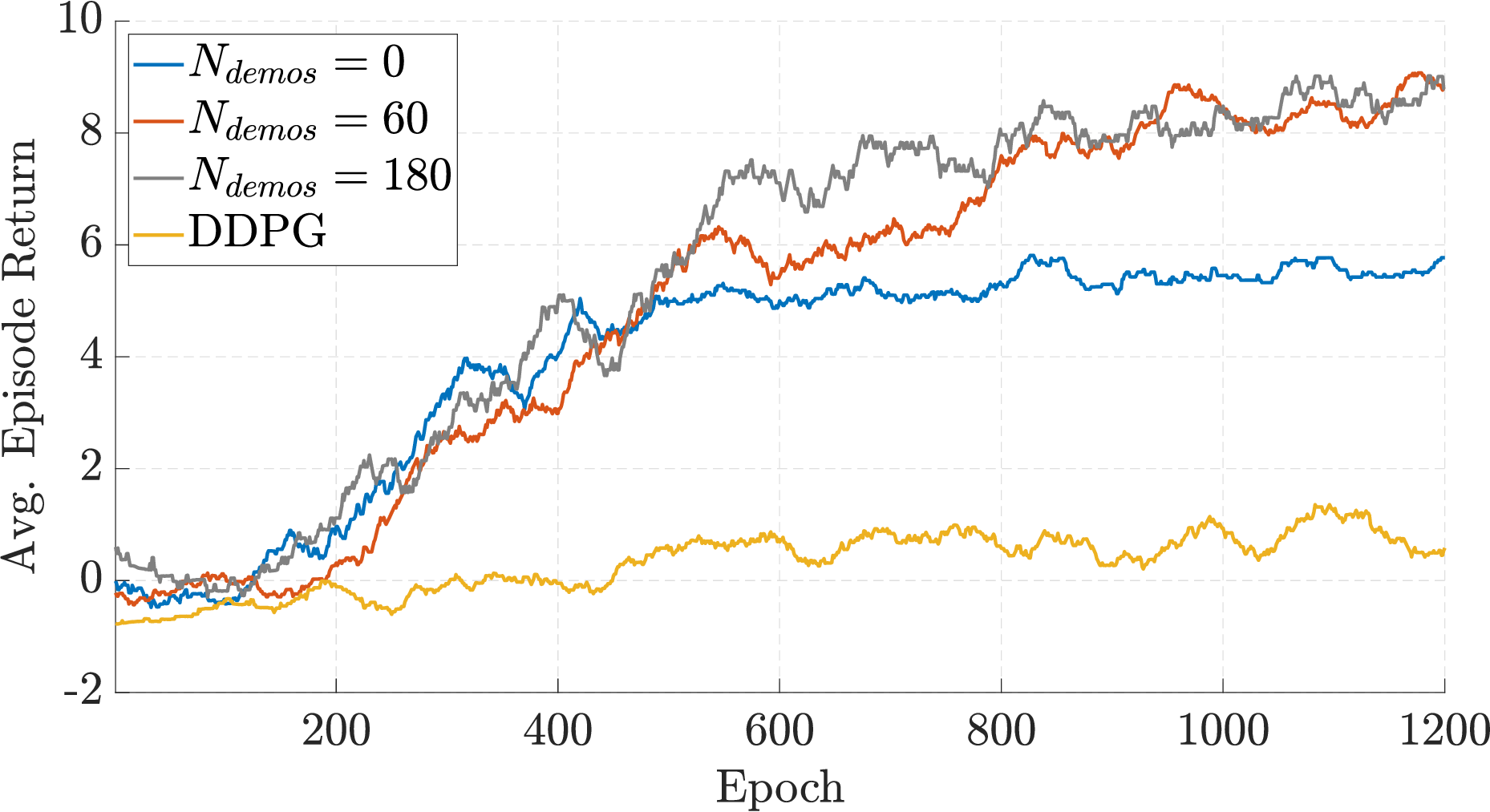}
        \caption{\small{Different numbers of demos $N_{\mathrm{demos}}$}}
        \label{fig:p2p_numdemo}
    \end{subfigure}
    \begin{subfigure}[b]{0.7\textwidth}
        \includegraphics[width=0.98\textwidth,clip]{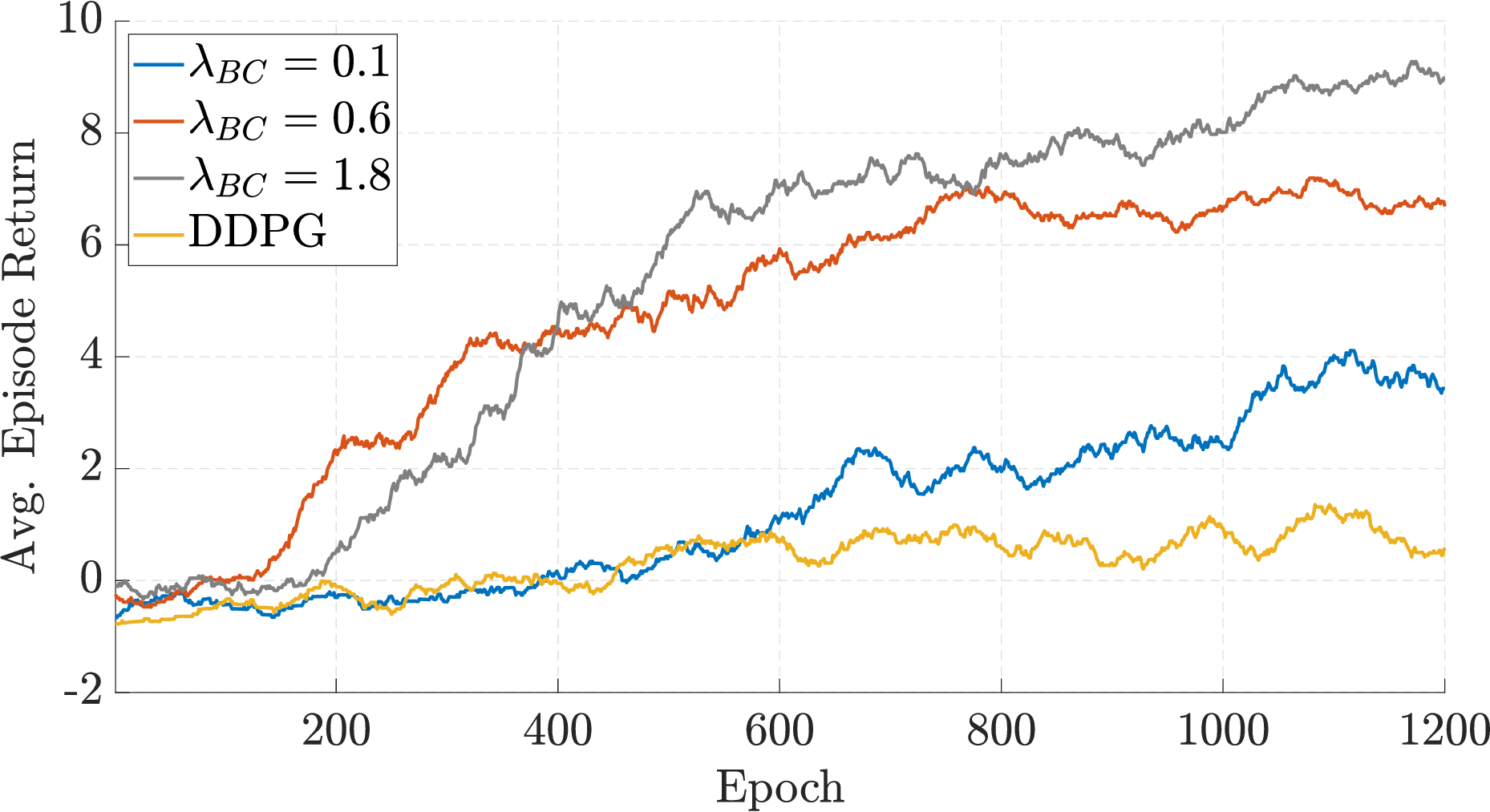}
        \caption{\small{Different behavior cloning gains $\lambda_{BC}$}}
        \label{fig:p2p_lambda}
    \end{subfigure}
    \caption{\small{The effect of hyperparameters on the P2P learning curves}}
    \label{fig:p2p_hyp}  
\end{figure}

The learning curves of the P2P task during the agent training stage are presented in Figure \ref{fig:p2p_hyp}. Specifically, the results depicting the influence of the number of demonstrations $N_{\mathrm{demos}}$ and the behavior cloning gain $\lambda_{BC}$ are respectively illustrated in Figure \ref{fig:p2p_numdemo} and Figure \ref{fig:p2p_lambda}. Additional training metrics are listed in Table \ref{tab:p2p}.

\begin{table}[htbp]
\begin{center}
\begin{minipage}{0.8\linewidth}
\caption{Numerical training metrics of the P2P agent}\label{tab:p2p}
\begin{tabular*}{0.98\linewidth}{@{\extracolsep{\fill}}lccccccc@{\extracolsep{\fill}}}
\toprule%
\multirow{2}{*}{Agent} & \multicolumn{3}{@{}c@{}}{$\lambda_{BC}$ ($N_{\mathrm{demos}} = 100$)} & \multicolumn{3}{@{}c@{}}{$N_{\mathrm{demos}}$ ($\lambda_{BC}=1$)} & \multirow{2}{*}{DDPG} \\
\cmidrule{2-4}\cmidrule{5-7}%
& 0.1 & 0.6 & 1.8 & 0 & 80 & 160 & \\
\midrule
$\overline{R}_{10}$ &-0.44	&-0.06	&-0.12	&-0.28	&-0.13	&0.06   &-0.56 \\
$\overline{R}_{90}$ &3.72	&6.83	&8.94   &5.55	&8.58	&8.67	&0.90
 \\
$I_R$ &4.16	&6.89	&9.06    &5.83	&8.71	&8.61	&1.49
 \\
$T_{50}$ &657	&292	&398	&281	&457	&379    &486 \\
\bottomrule
\end{tabular*}
\end{minipage}
\end{center}
\end{table}

\subsubsection{The P2P-O Task}
\begin{figure}[hbtp]
\centering
    \begin{subfigure}[b]{0.7\textwidth}
        \centering
        \includegraphics[width=0.98\textwidth,clip]{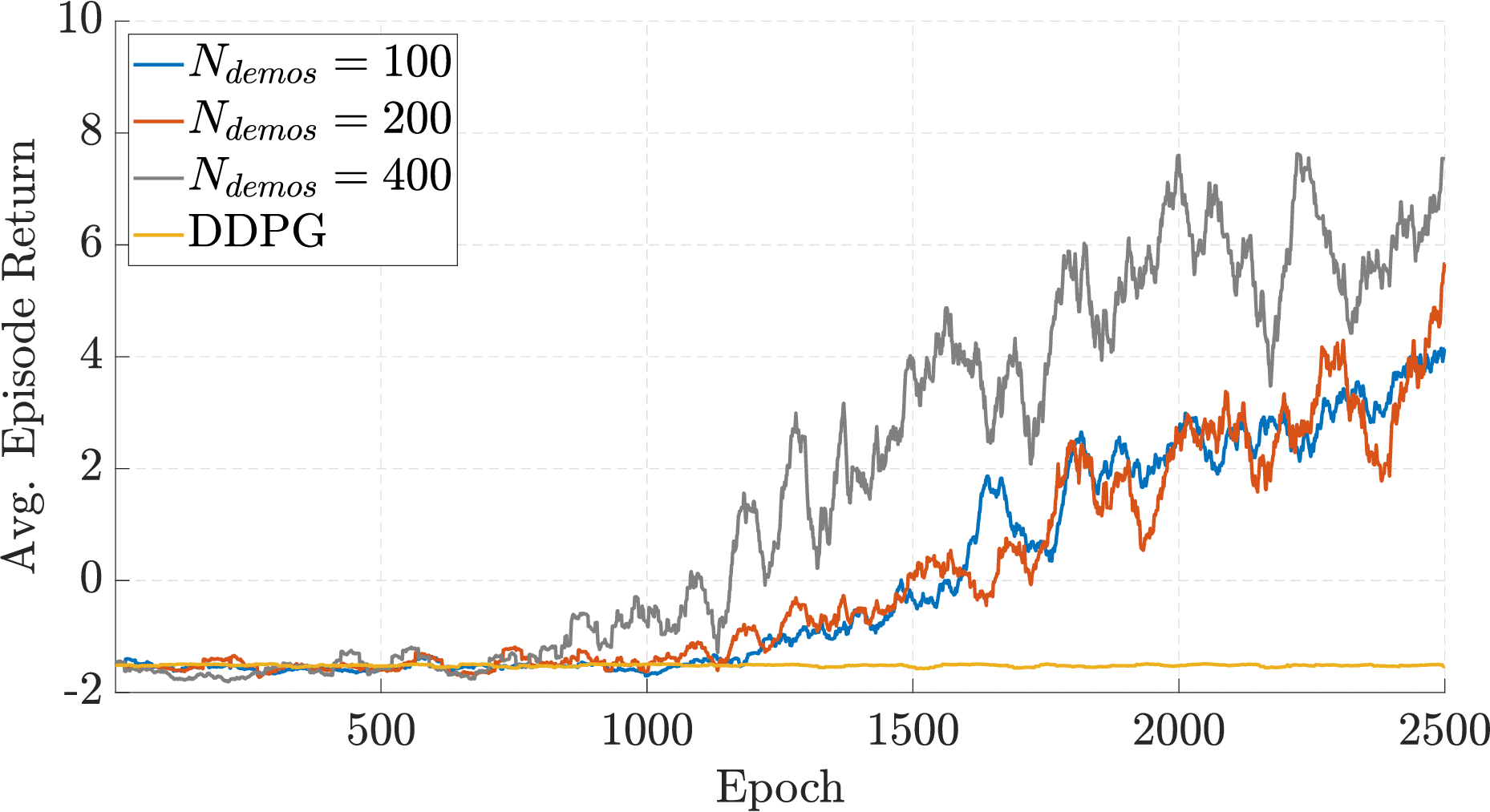}
        \caption{\small{Different numbers of demos $N_{\mathrm{demos}}$}}
        \label{fig:p2p_obs_numdemo}
    \end{subfigure}
    \begin{subfigure}[b]{0.7\textwidth}
        \centering
        \includegraphics[width=0.98\textwidth,clip]{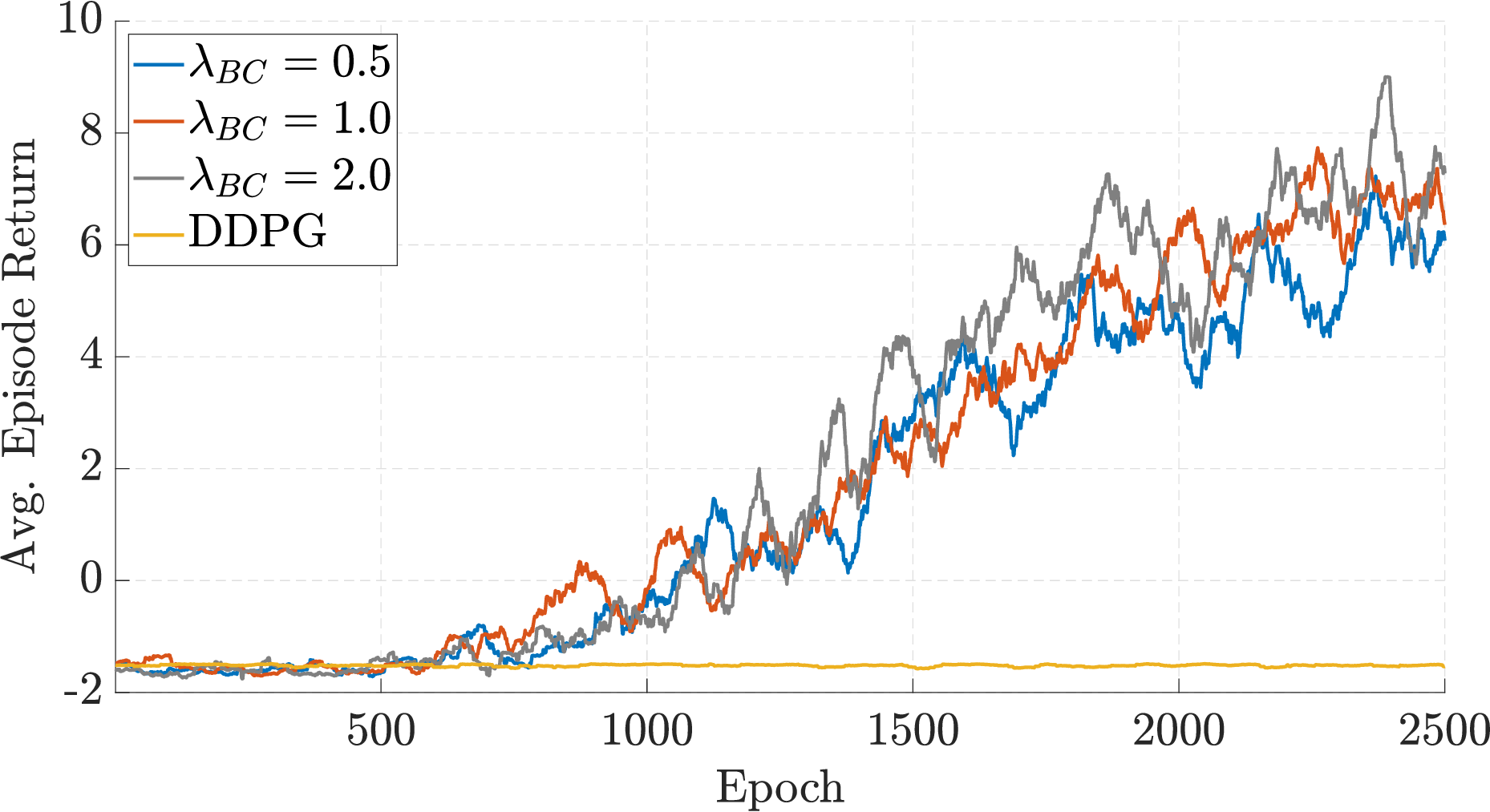}
        \caption{\small{Different behavior cloning gains $\lambda_{BC}$}}
        \label{fig:p2p_obs_lambda}
    \end{subfigure}
    \caption{\small{The effect of hyperparameters on the P2P-O learning curves}}
    \label{fig:p2p_obs_hyp}
\end{figure}

The learning curves of the P2P-O task during the agent training stage are shown in Figure \ref{fig:p2p_obs_hyp}. Specifically, the results showing the influence of the number of demonstrations $N_{\mathrm{demos}}$ and the behavior cloning gain $\lambda_{BC}$ are respectively illustrated in Figure \ref{fig:p2p_obs_numdemo} and Figure \ref{fig:p2p_obs_lambda}. Other numerical metrics are shown in Table \ref{tab:p2p-o}.

\begin{table}[htbp]
\begin{center}
\begin{minipage}{0.8\linewidth}
\caption{Numerical training metrics of the P2P-O agent}\label{tab:p2p-o}
\begin{tabular*}{0.96\linewidth}{@{\extracolsep{\fill}}lccccccc@{\extracolsep{\fill}}}
\toprule%
\multirow{2}{*}{Agent} & \multicolumn{3}{@{}c@{}}{$\lambda_{BC}$ ($N_{\mathrm{demos}} = 250$)} & \multicolumn{3}{@{}c@{}}{$N_{\mathrm{demos}}$ ($\lambda_{BC}=1$)} & \multirow{2}{*}{DDPG} \\
\cmidrule{2-4}\cmidrule{5-7}%
& 0.5 & 1 & 2 & 100 & 200 & 400 & \\
\midrule
$\overline{R}_{10}$ &-1.61	&-1.51	&-1.62   &-1.53	&-1.51	&-1.67	&-1.51 \\
$\overline{R}_{90}$ &5.79	&6.80	&7.19    &3.40	&3.44	&5.95	&-1.52 \\
$I_R$ &7.40     &8.31	&8.81    &4.93	&4.95	&7.63	&-0.01 \\
$T_{50}$ &1503	&1603	&1441    &1787	&1780	&1483	&N/A \\
\bottomrule
\end{tabular*}
\end{minipage}
\end{center}
\end{table}

We have selected the hyperparameters, $\lambda_{BC}$ and $N_{\mathrm{demos}}$ values in Tables~\labelcref{tab:pnp,tab:p2p,tab:p2p-o} and Figures~\labelcref{fig:pnp_hyp,fig:p2p_hyp,fig:p2p_obs_hyp}, such that they provide a clear performance advantage, starting with values that enable a minimal partial improvement and then progressively increasing them until the effect of demonstrations saturates.
{It could be observed that the range of the hyperparameters are subject to change based on the environment complexity and algotihm requirement. Therefore, some level of finetuning might still be needed, but the demonstration data requirement in all cases is relatively low compared to the original data requirements of RL.}
This approach helps illustrate both the low-end sensitivity and diminishing returns at higher values. In the P\&P and P2P tasks, this trend is evident in the learning curves and metrics. For P2P-O, however, the masking mechanism acts more like a binary switch. Nearly all effective hyperparameter values produce comparable outcomes, reflecting lower sensitivity in this environment. We chose the values in Figures~\labelcref{fig:pnp_hyp,fig:p2p_hyp} accordingly to reflect meaningful variation where applicable.

\subsection{Test Study}
In this section, we deploy the trained agents onto the robot simulation model and evaluate their test performance. The test study is conducted in the global environment, with 500 trials per task. In the test study, we use the following metrics for evaluation. 
\begin{itemize}
\item \textbf{Success rate} $P_{\mathrm{scs}}$: the ratio of the number of successful trials $N_{\mathrm{scs}}$ to all trials $N_{\mathrm{ttl}}$.
\item \textbf{Average effort} $\overline{T}_{\mathrm{eff}}$: average torque utility rate of the joints in the successful trials, as in the following,
\begin{equation}
    \overline{T}_{\mathrm{eff}} = \frac{1}{N_{\mathrm{scs}}} \frac{1}{T} \sum_{i=1}^{N_{\mathrm{scs}}} \sum_{t=1}^{T}  \left\| \hat{\tau}_t^i \right\|.
\end{equation}
where $\hat{\tau}_t$ is the norm of the vector of active joint utilities, and $i$ denotes the $i$-th trial. 

\item \textbf{Average test return} $\overline{R}_{\mathrm{test}}$: the average return over successful trials.
 
\item \textbf{Ultimate error} $\overline{E}_{\overline{95}}$: the average ultimate reaching error over the final 5\% of a successful trial,
\begin{equation}
\overline{E}_{\overline{95}} = \frac{1}{N_{\mathrm{scs}}} \frac{1}{0.05T} \sum_{j=1}^{N_{\mathrm{scs}}} \sum_{t=0.95T}^{T} \left\|e_t^{(j)} \right\|,
\end{equation}
where $e_t^{(j)}$ is the reaching error of trial $j$ at time $t$.
\end{itemize}

\begin{figure}[htbp]
\centering
    \begin{subfigure}[b]{0.48\textwidth}
        \centering
        \includegraphics[width=0.95\textwidth,clip]{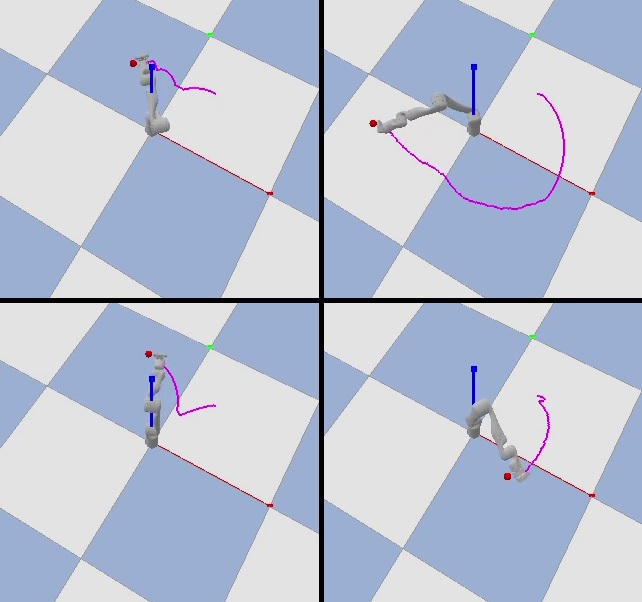}
        \caption{P2P}
        \label{fig:p2p_ease}
    \end{subfigure}
    \begin{subfigure}{0.48\textwidth}
        \centering
        \includegraphics[width=0.95\textwidth,clip]{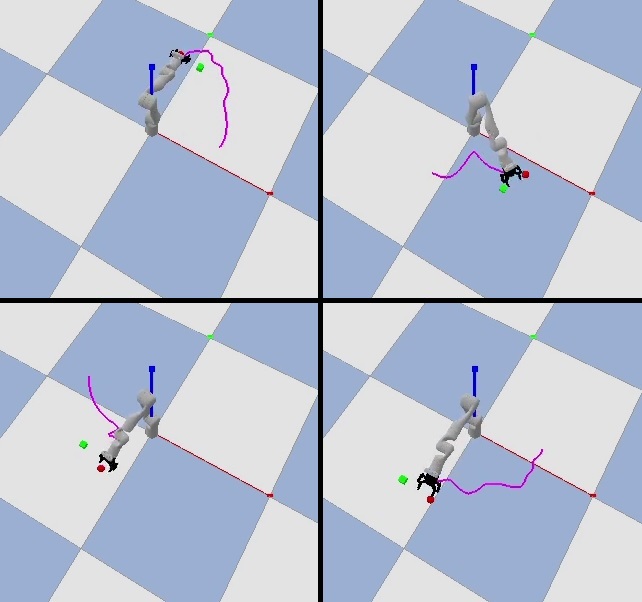}
        \caption{P2P with obstacle}
        \label{fig:p2p_obs_ease}
    \end{subfigure}
    \caption{\small{Samples of successful trials of trained agents}}
    \label{fig:ease}
\end{figure}

The initial robot end-effector position is fixed as $\theta_0=-{\pi}/{4}$, $\rho_0=0.5$, $z_0=0.45$ for P2P and $\theta_0=-{\pi}/{2}$, $\rho_0=0.5$, $z_0=0.45$ for P2P-O and P\&P. The robot goal configuration is sampled from the same distribution as the training configuration in Section~\ref{sec:def_reward}. Each test trial ends if the \textbf{Reaching}, \textbf{Timeout}, and \textbf{Collision} criteria are met. Besides, we refer to the trajectories that meet the \textbf{Reaching} criterion as \textit{successful trials}. Examples of the successful trials in two of the simulation environments of the test study are illustrated in Figure \ref{fig:ease}.

\begin{figure}[htbp]
\centering
    \begin{subfigure}[b]{0.48\textwidth}
        \centering
        \includegraphics[width=0.98\textwidth,clip]{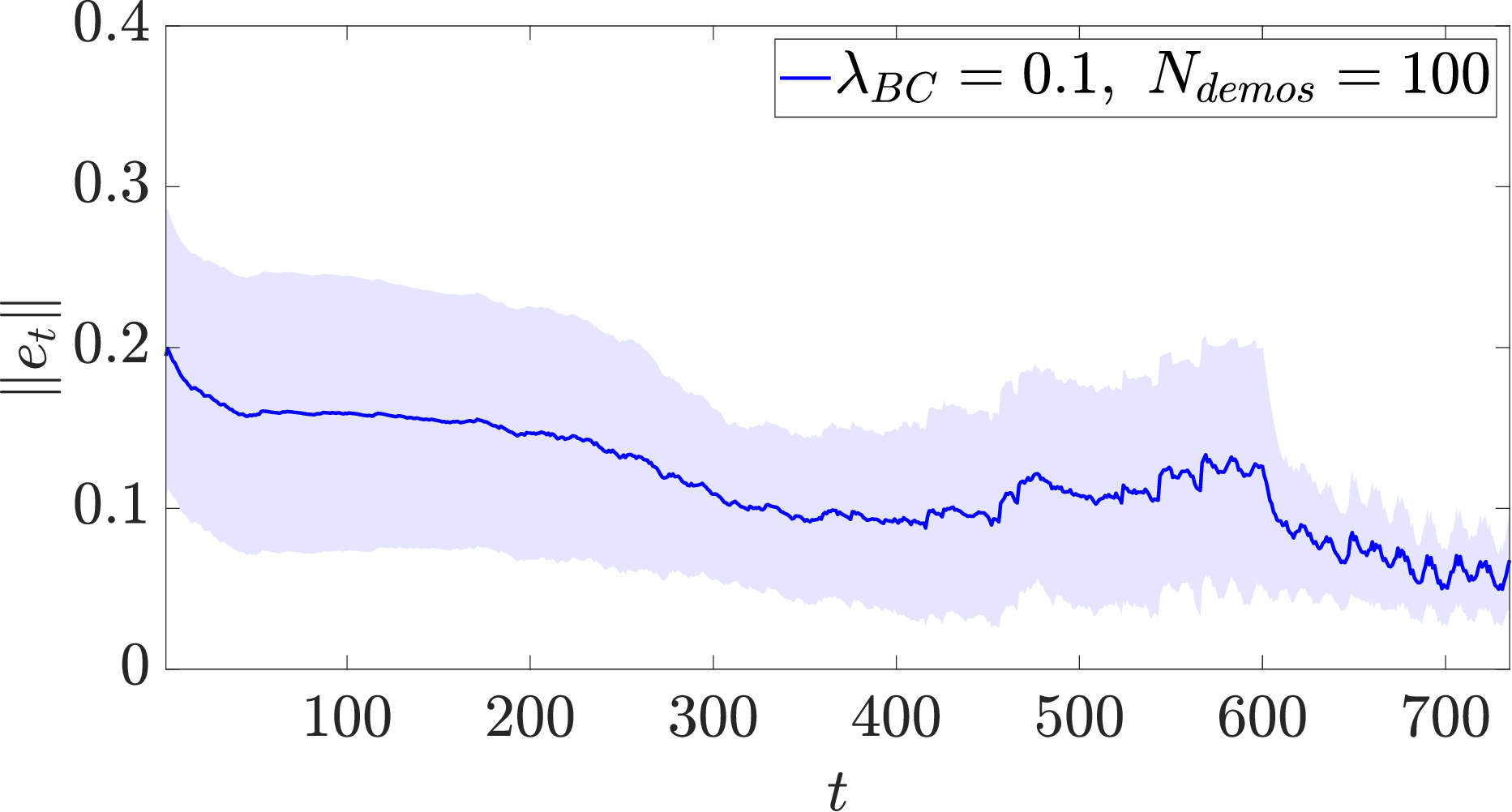}
    \end{subfigure}
    \begin{subfigure}[b]{0.48\textwidth}
        \centering
        \includegraphics[width=0.98\textwidth,clip]{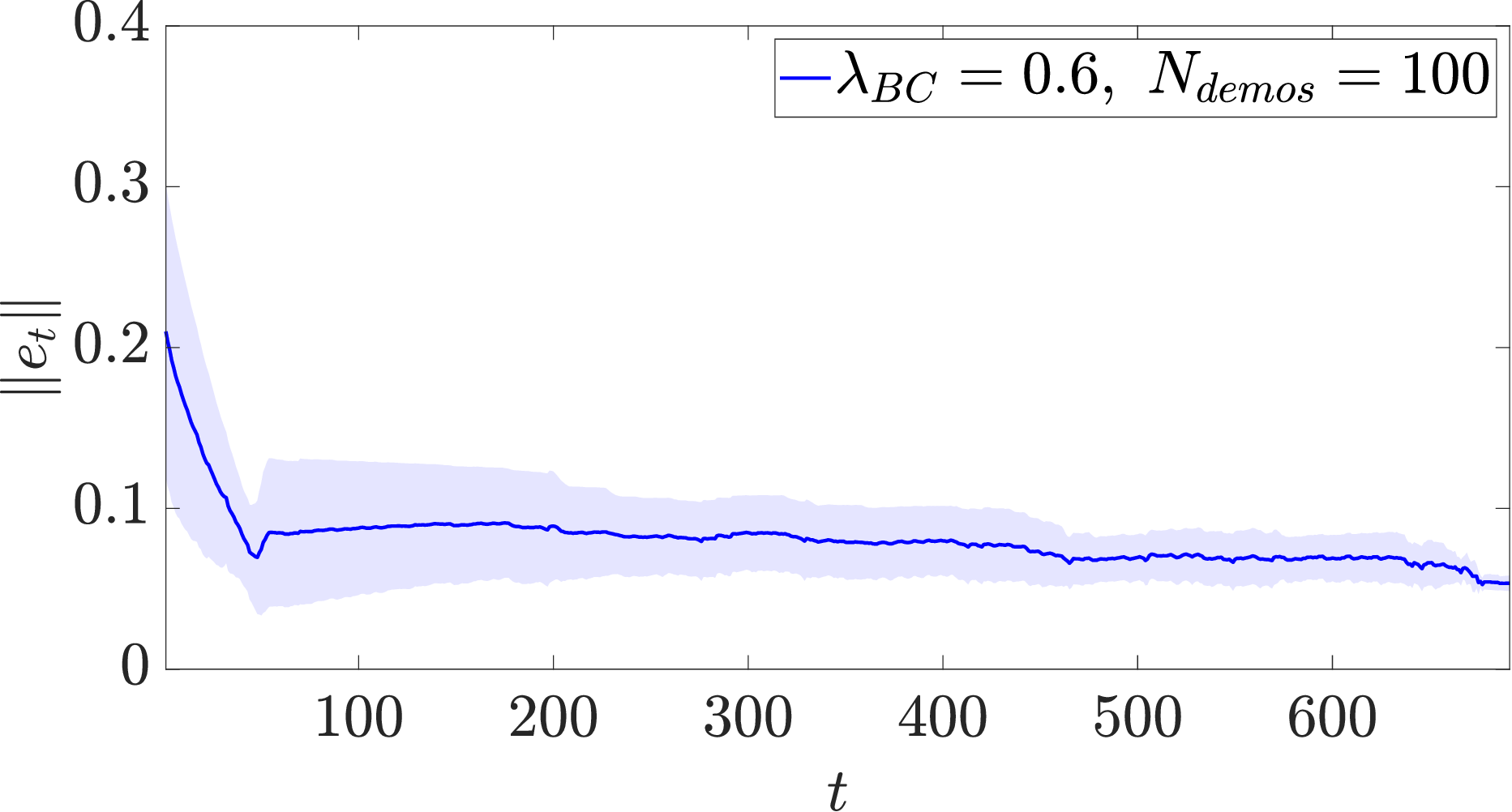}
    \end{subfigure}
    \begin{subfigure}[b]{0.48\textwidth}
        \centering
        \includegraphics[width=0.98\textwidth,clip]{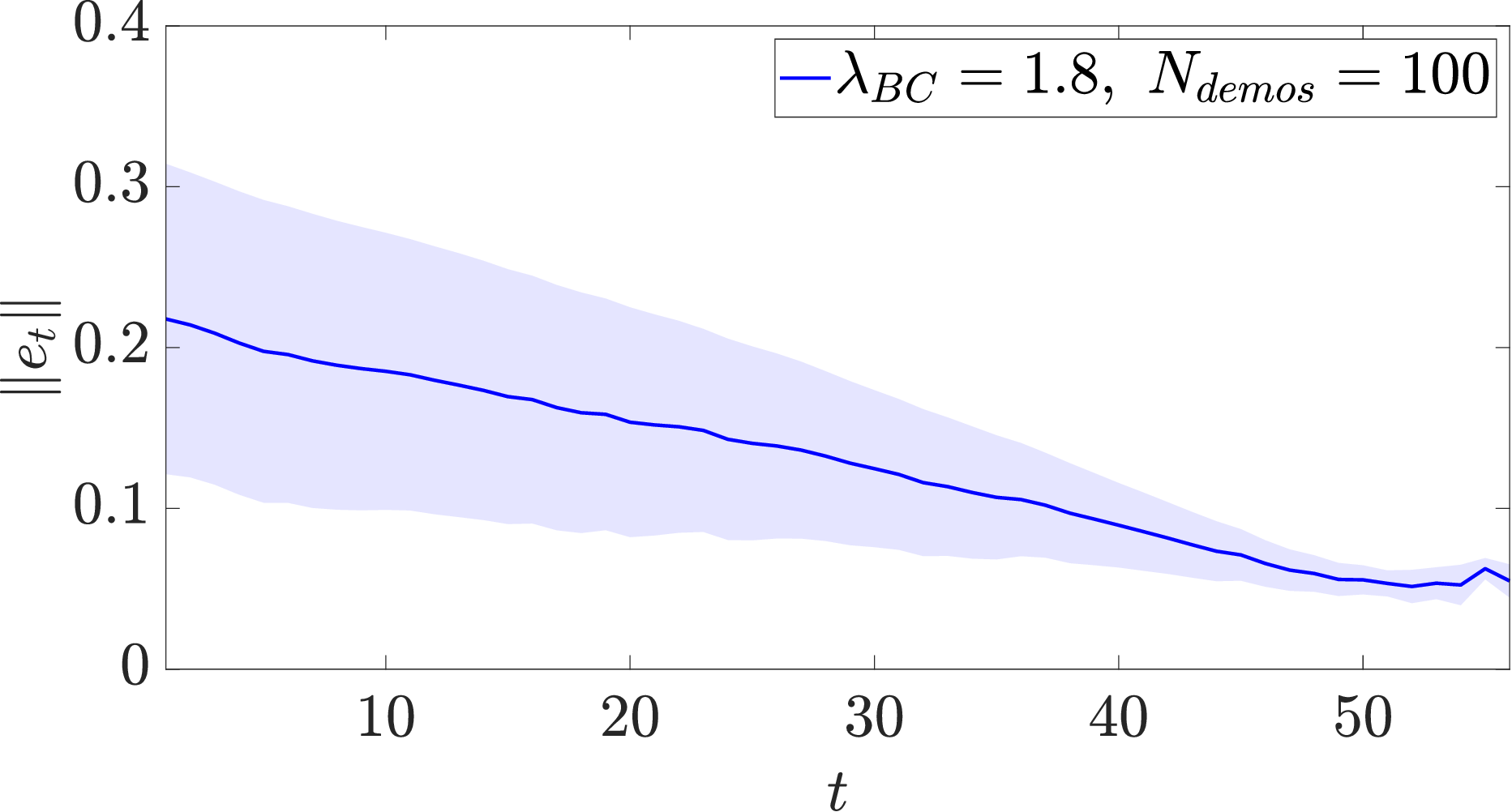}
    \end{subfigure}
    \begin{subfigure}[b]{0.48\textwidth}
        \centering
        \includegraphics[width=0.98\textwidth,clip]{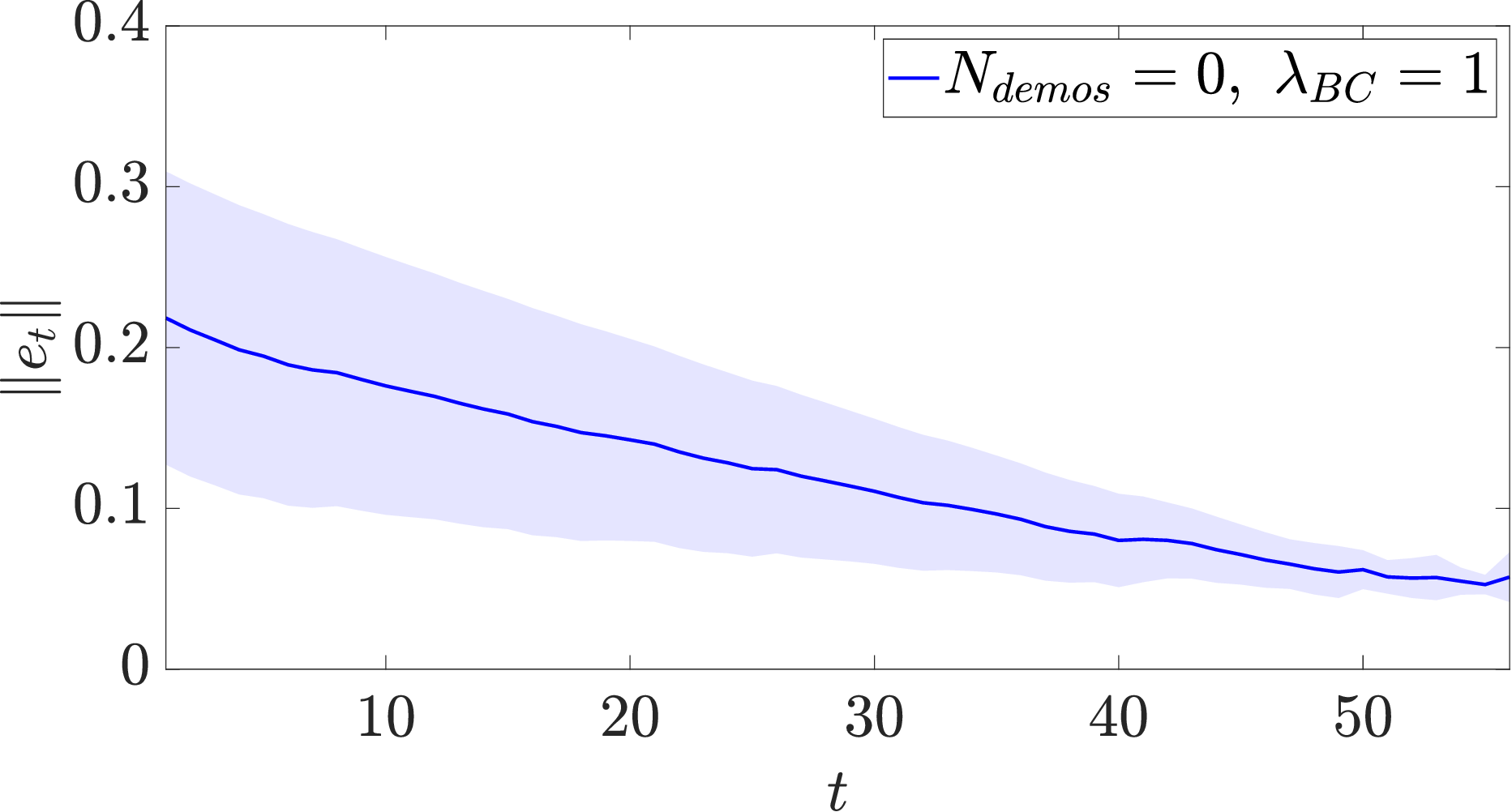}
    \end{subfigure}
    \begin{subfigure}[b]{0.48\textwidth}
        \centering
        \includegraphics[width=0.98\textwidth,clip]{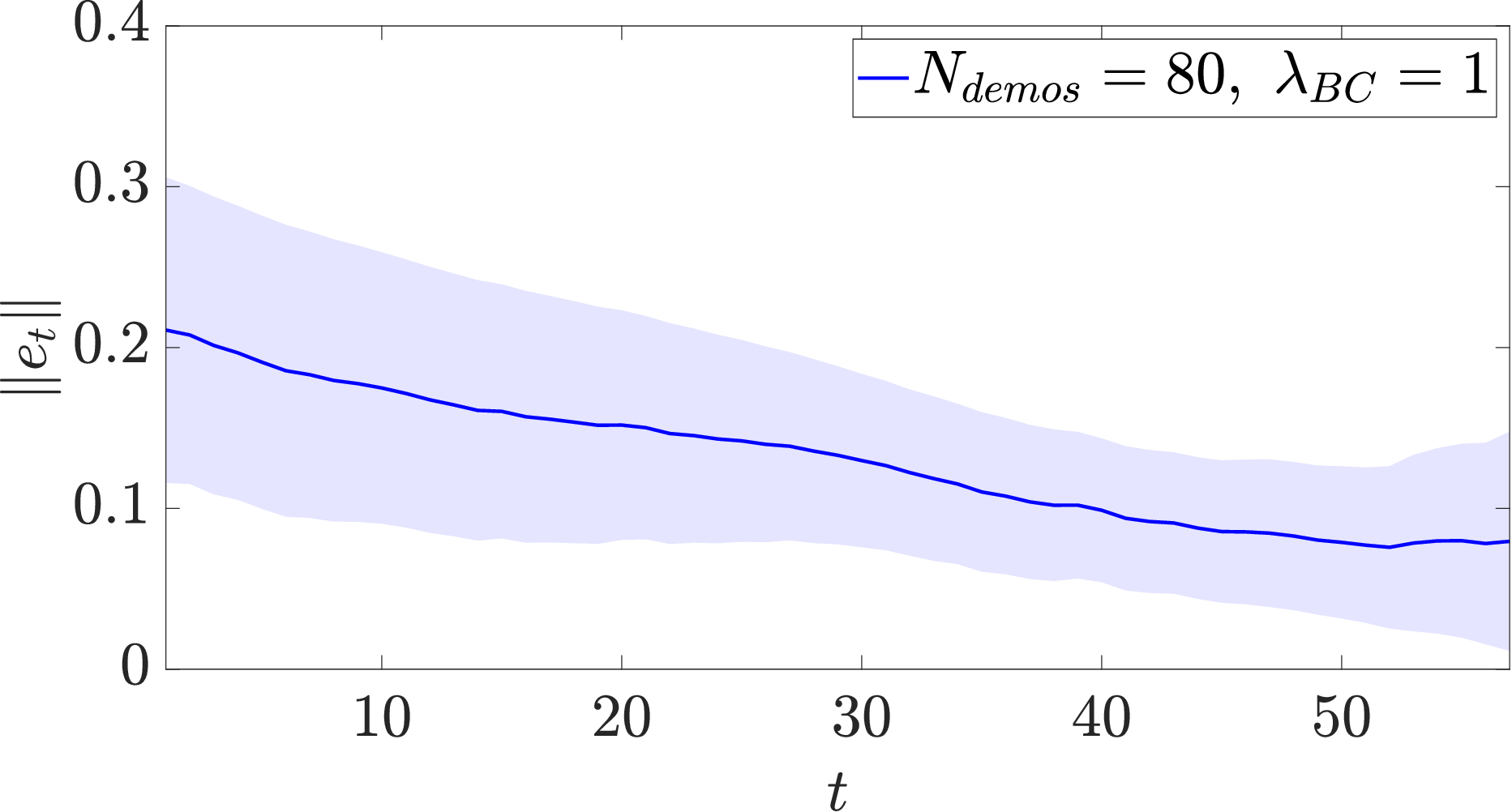}
    \end{subfigure}
    \begin{subfigure}[b]{0.48\textwidth}
        \centering
        \includegraphics[width=0.98\textwidth,clip]{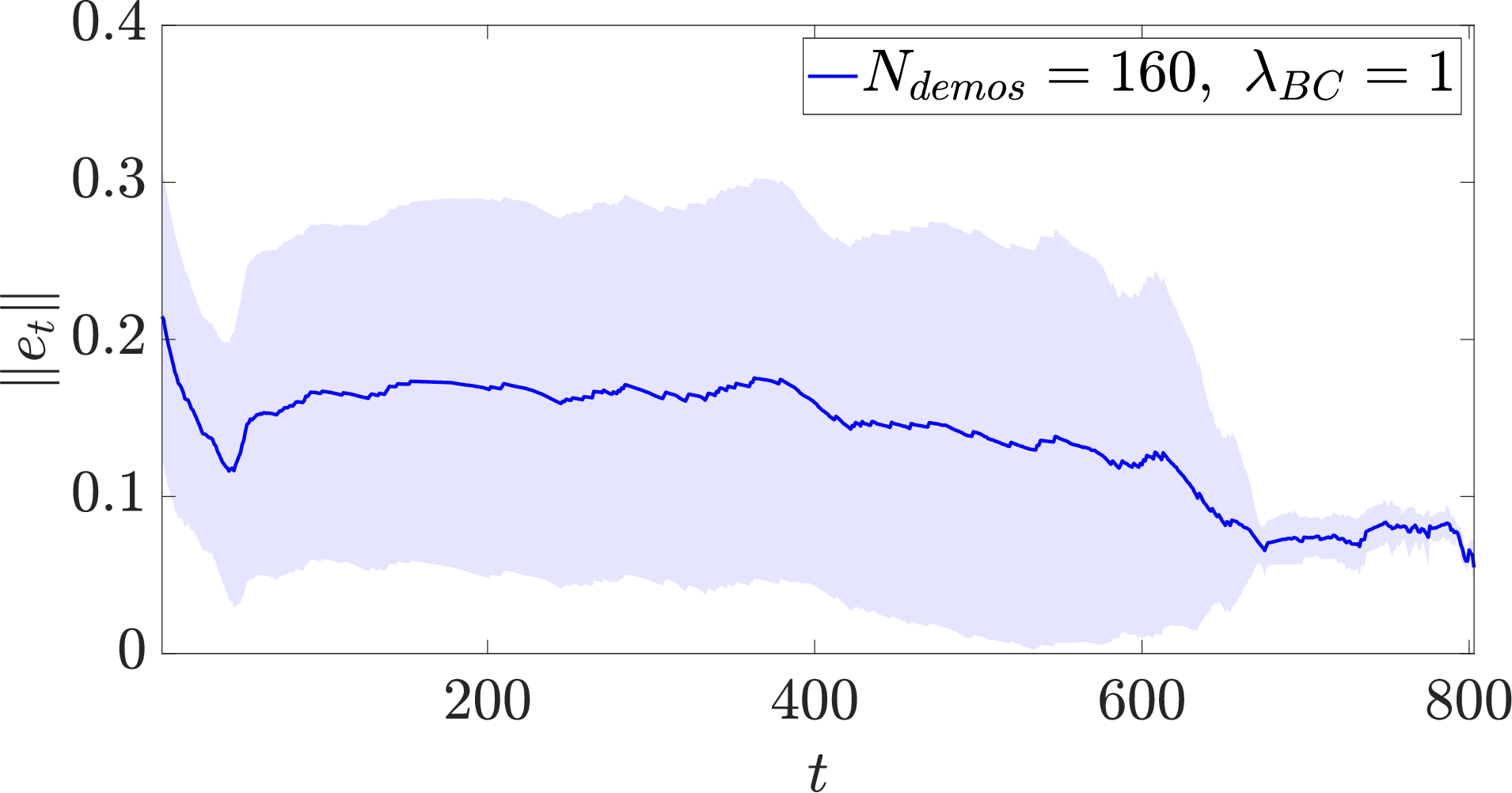}
    \end{subfigure}
    \caption{The average error $\|e_t\|$ in successful trials over time $t$ in the P2P test.}
    \label{fig:p2p_errors}
\end{figure}

\begin{figure}[htbp]
\centering
    \begin{subfigure}[b]{0.48\textwidth}
        \includegraphics[width=0.98\textwidth,clip]{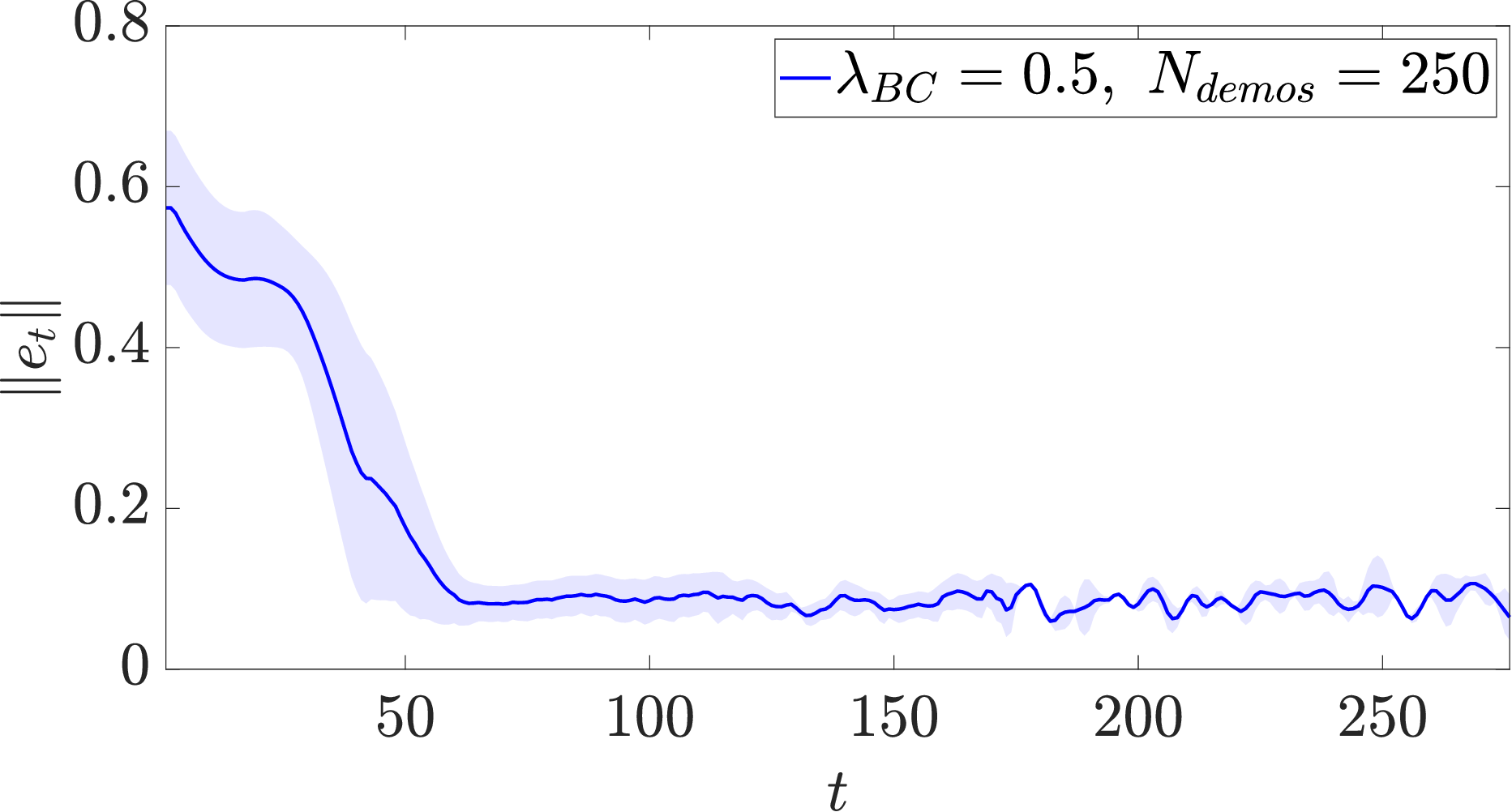}
    \end{subfigure}
    \begin{subfigure}[b]{0.48\textwidth}
        \includegraphics[width=0.98\textwidth,clip]{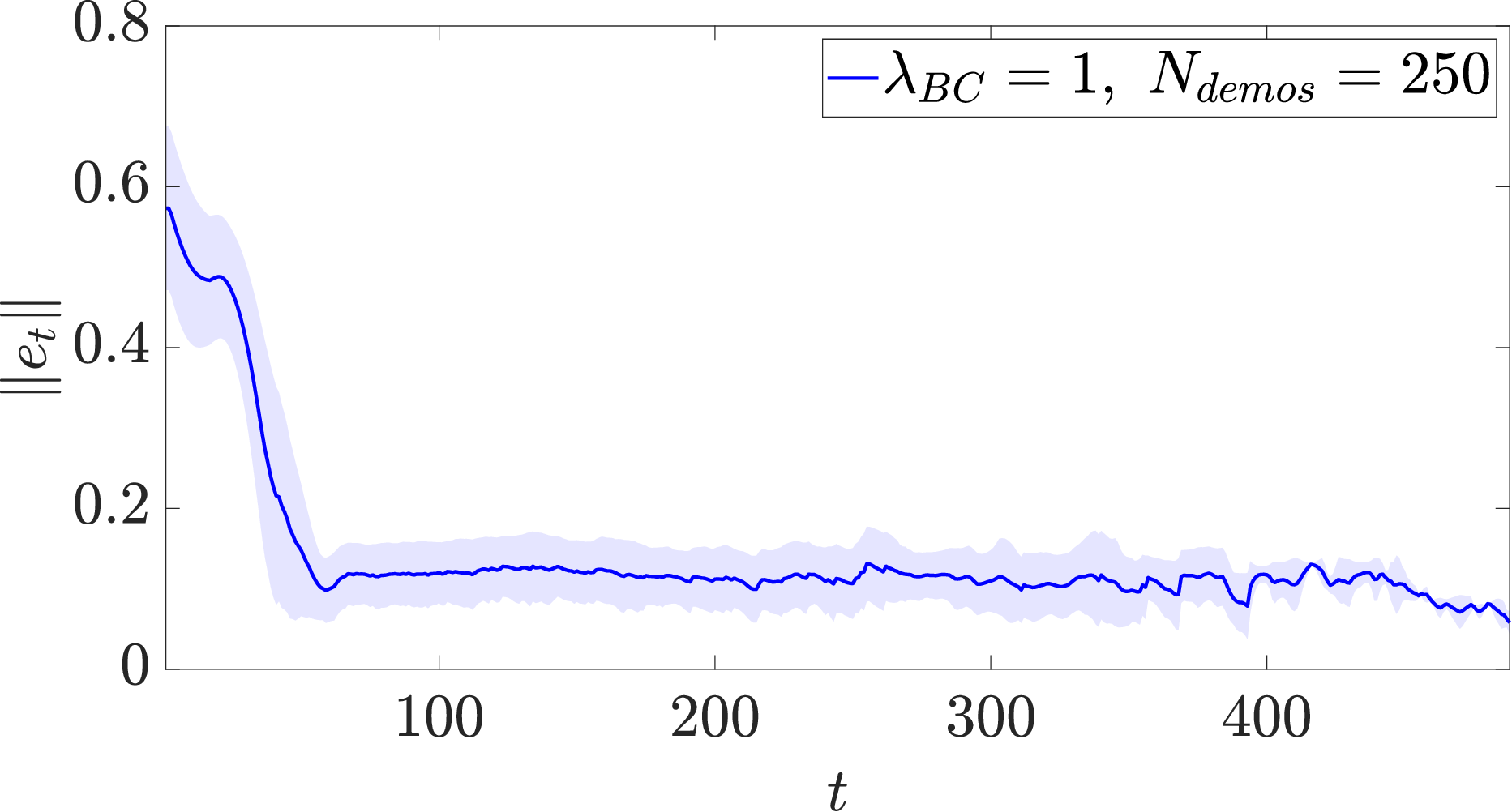}
    \end{subfigure}
    \begin{subfigure}[b]{0.48\textwidth}
        \includegraphics[width=0.98\textwidth,clip]{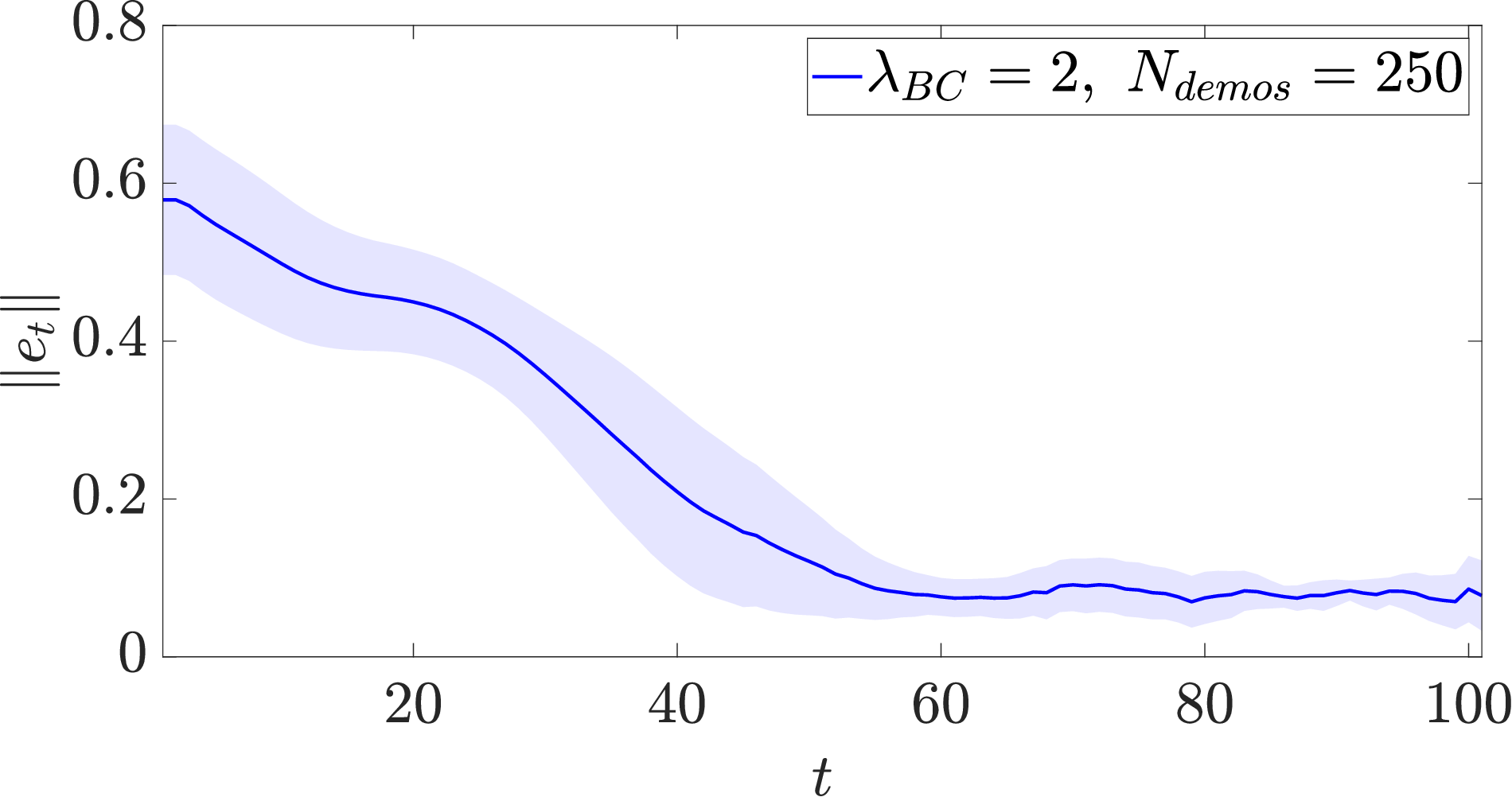}
    \end{subfigure}
    \begin{subfigure}[b]{0.48\textwidth}
        \includegraphics[width=0.98\textwidth,clip]{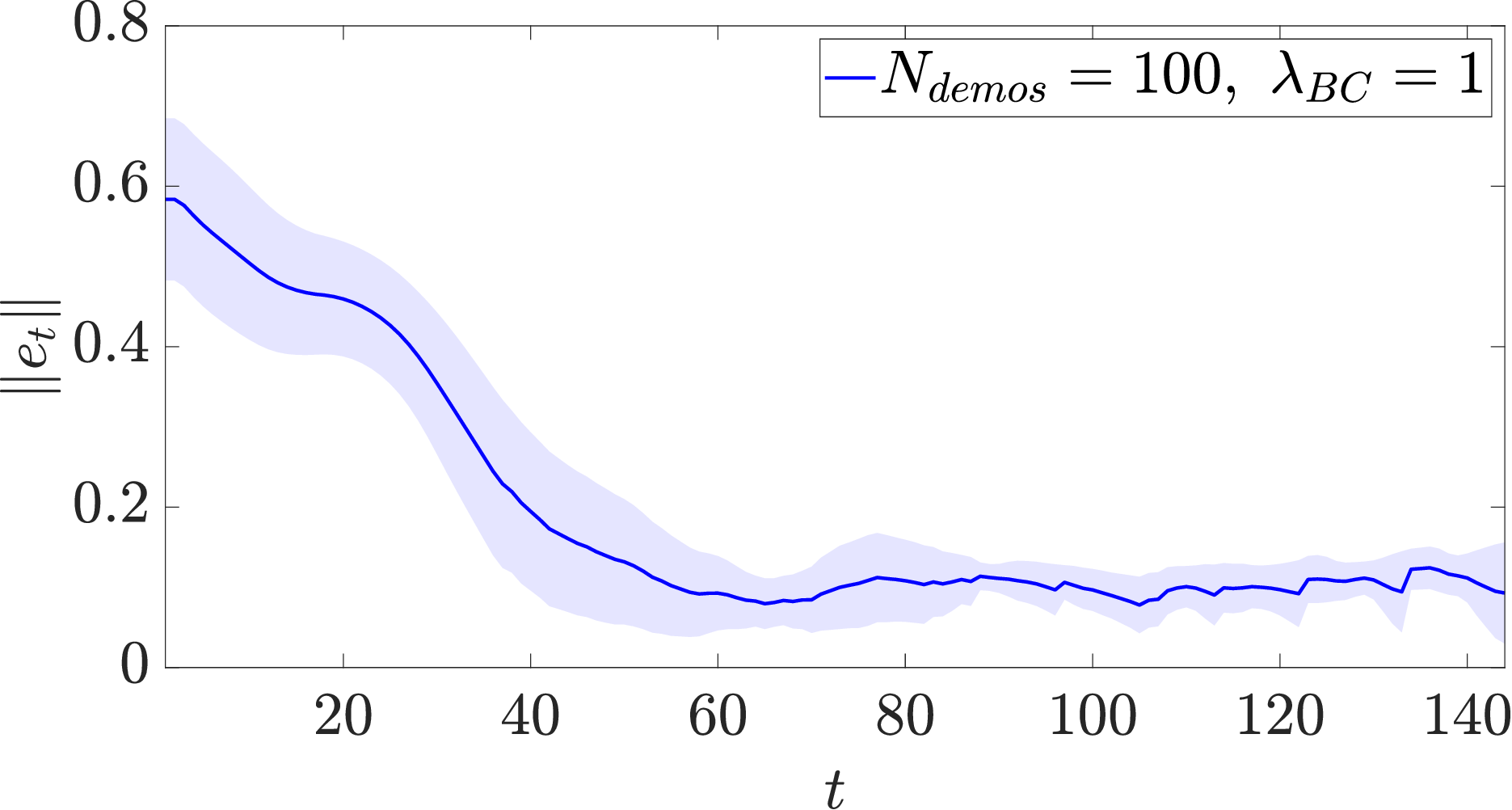}
    \end{subfigure}
    \begin{subfigure}[b]{0.48\textwidth}
        \includegraphics[width=0.98\textwidth,clip]{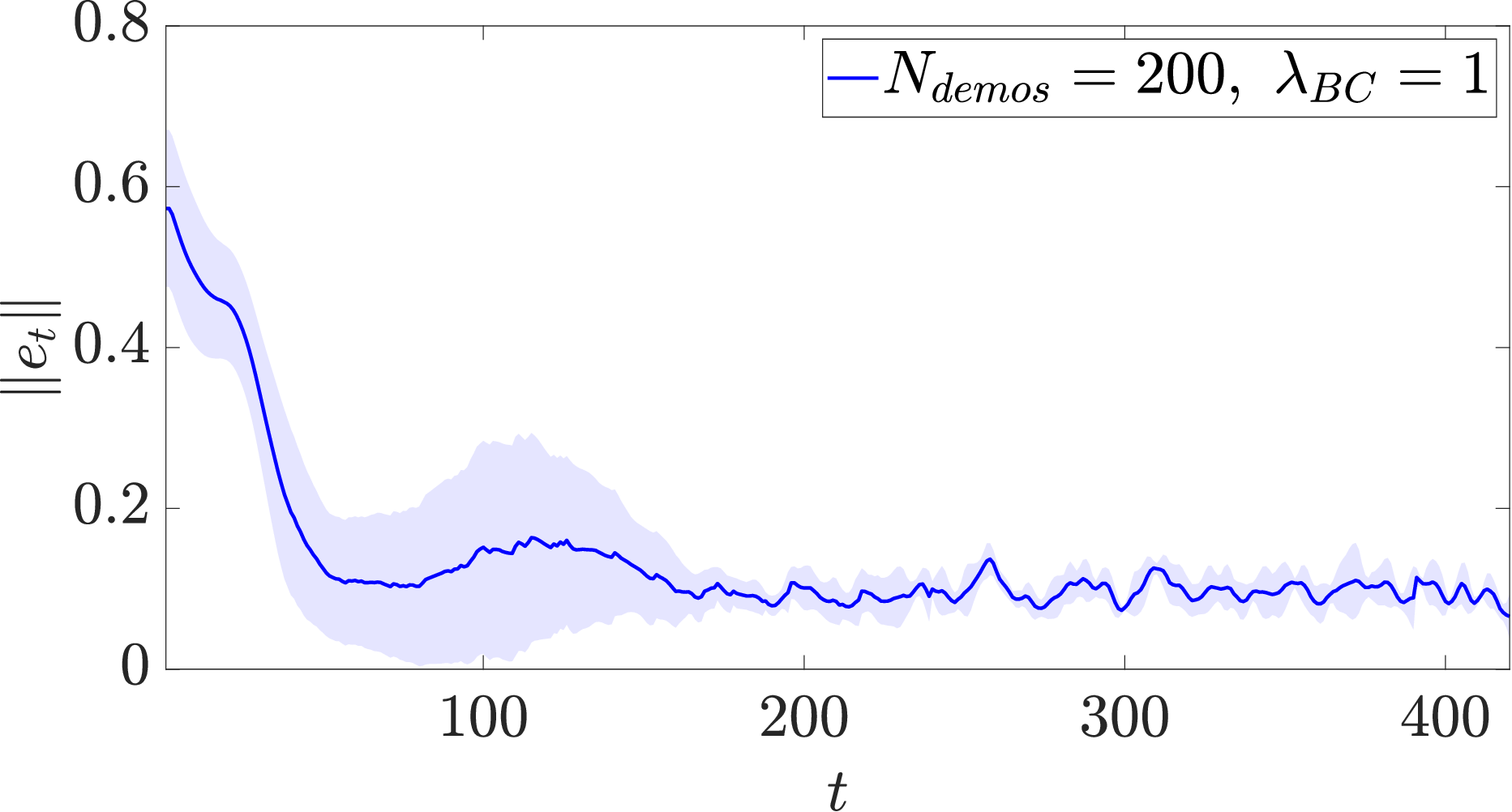}
    \end{subfigure}
    \begin{subfigure}[b]{0.48\textwidth}
        \includegraphics[width=0.98\textwidth,clip]{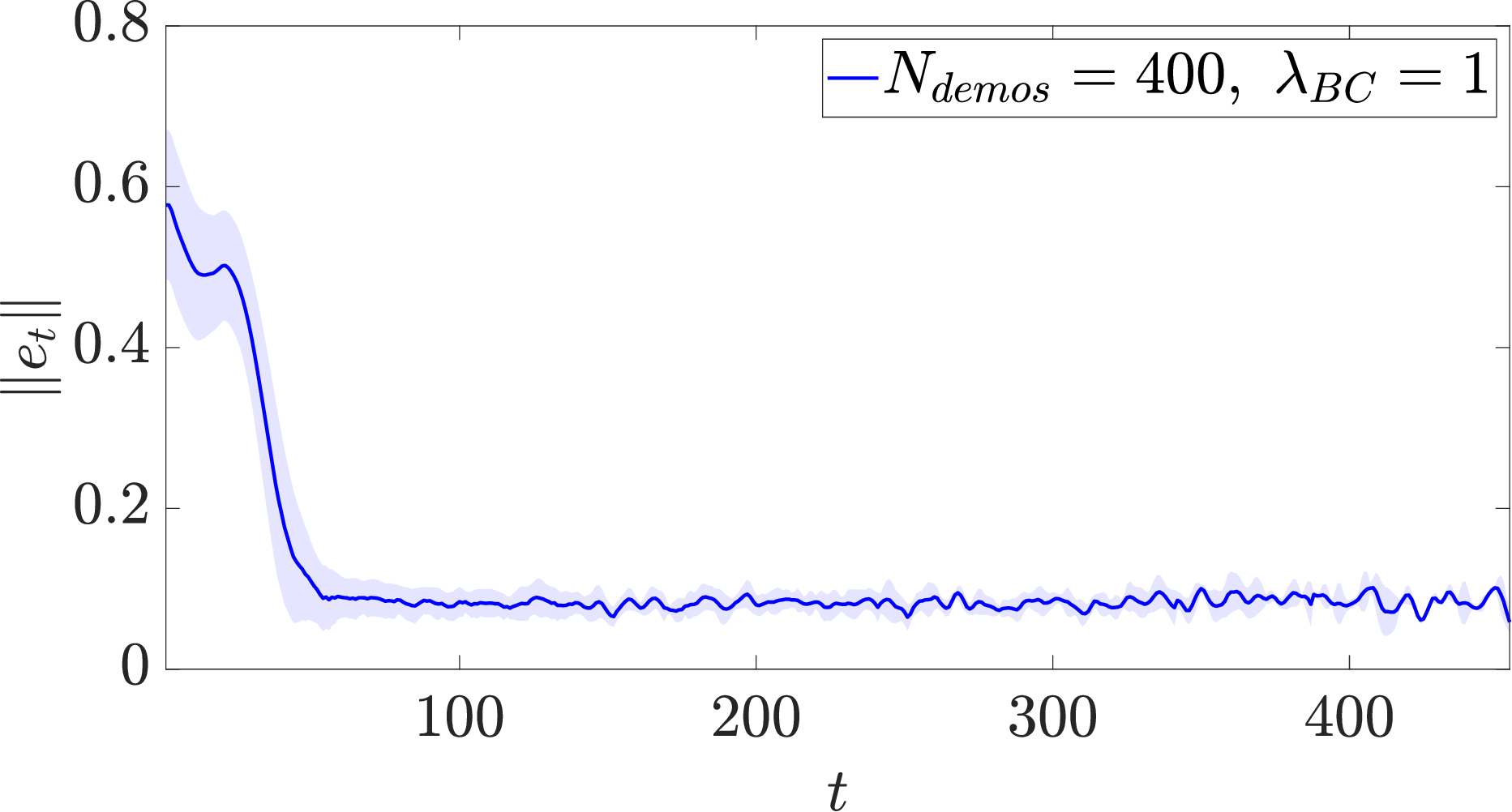}
    \end{subfigure}
    \caption{The average error $\|e_t\|$ and in successful trials over time $t$ in the P2P-O test.}
    \label{fig:p2po_errors}
\end{figure}

\begin{figure}[htbp]
\centering
    \begin{subfigure}[b]{0.48\textwidth}
        \includegraphics[width=0.98\textwidth,clip]{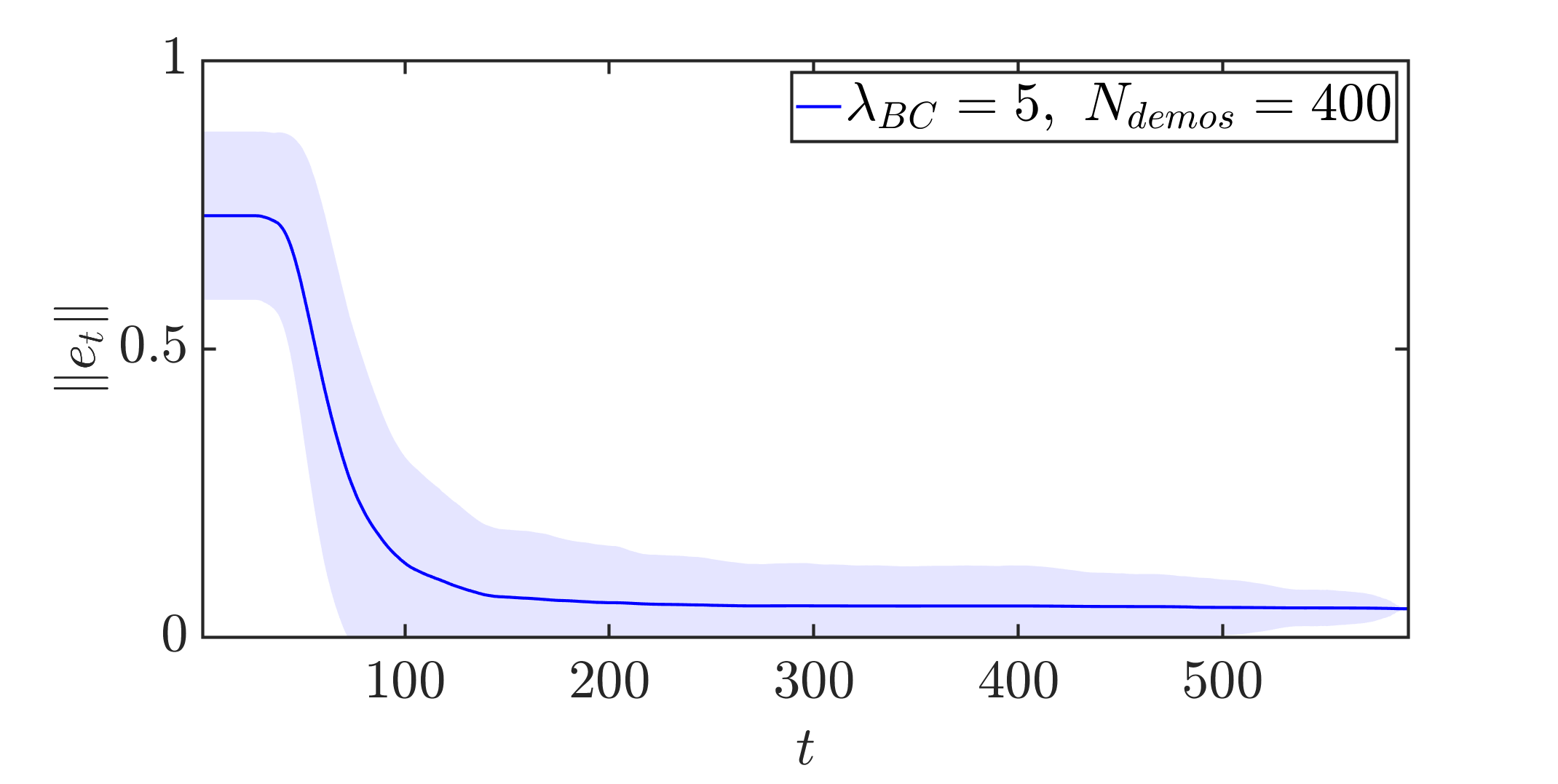}
    \end{subfigure}
    \begin{subfigure}[b]{0.48\textwidth}
        \includegraphics[width=0.98\textwidth,clip]{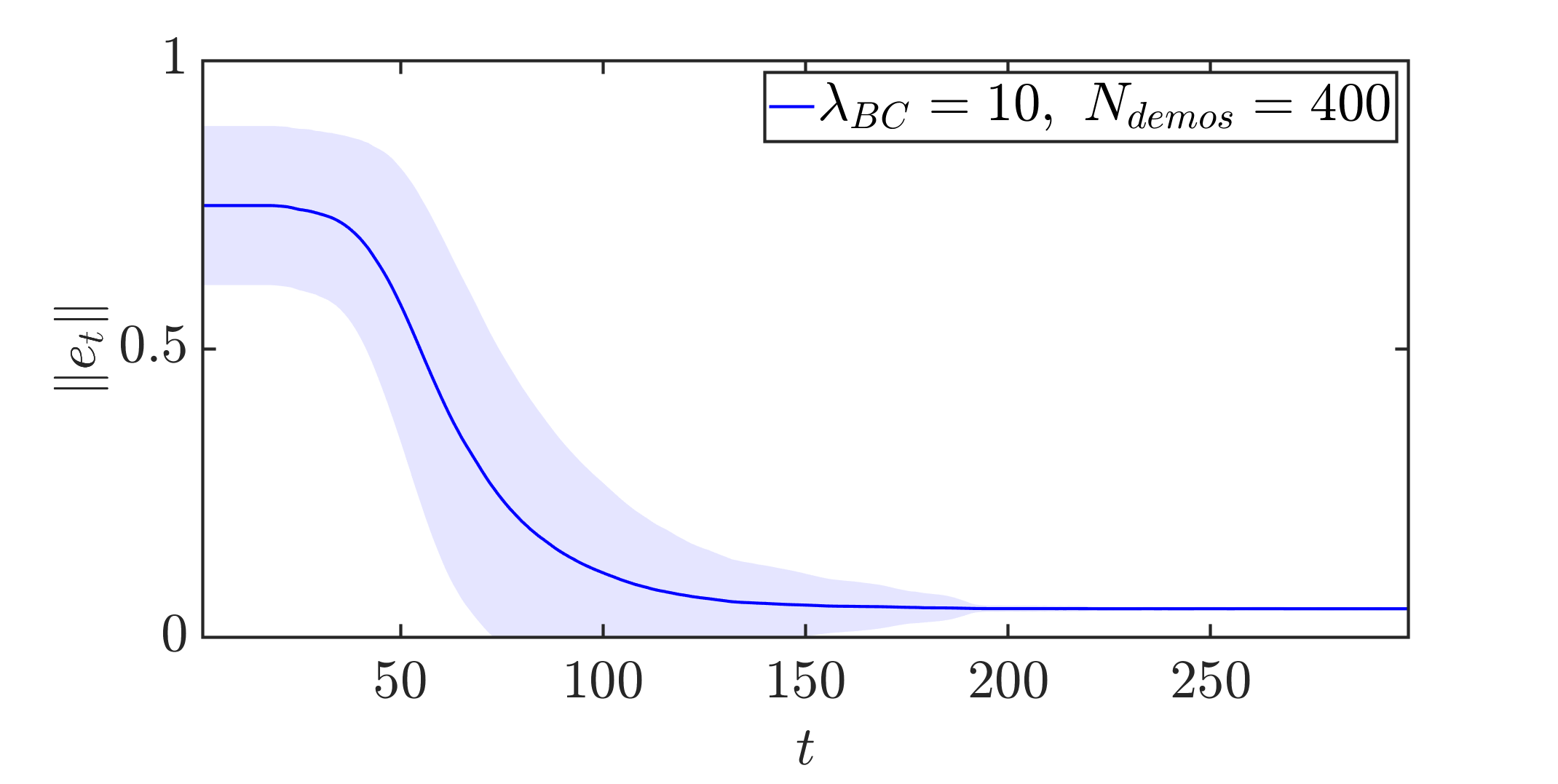}
    \end{subfigure}
    \begin{subfigure}[b]{0.48\textwidth}
        \includegraphics[width=0.98\textwidth,clip]{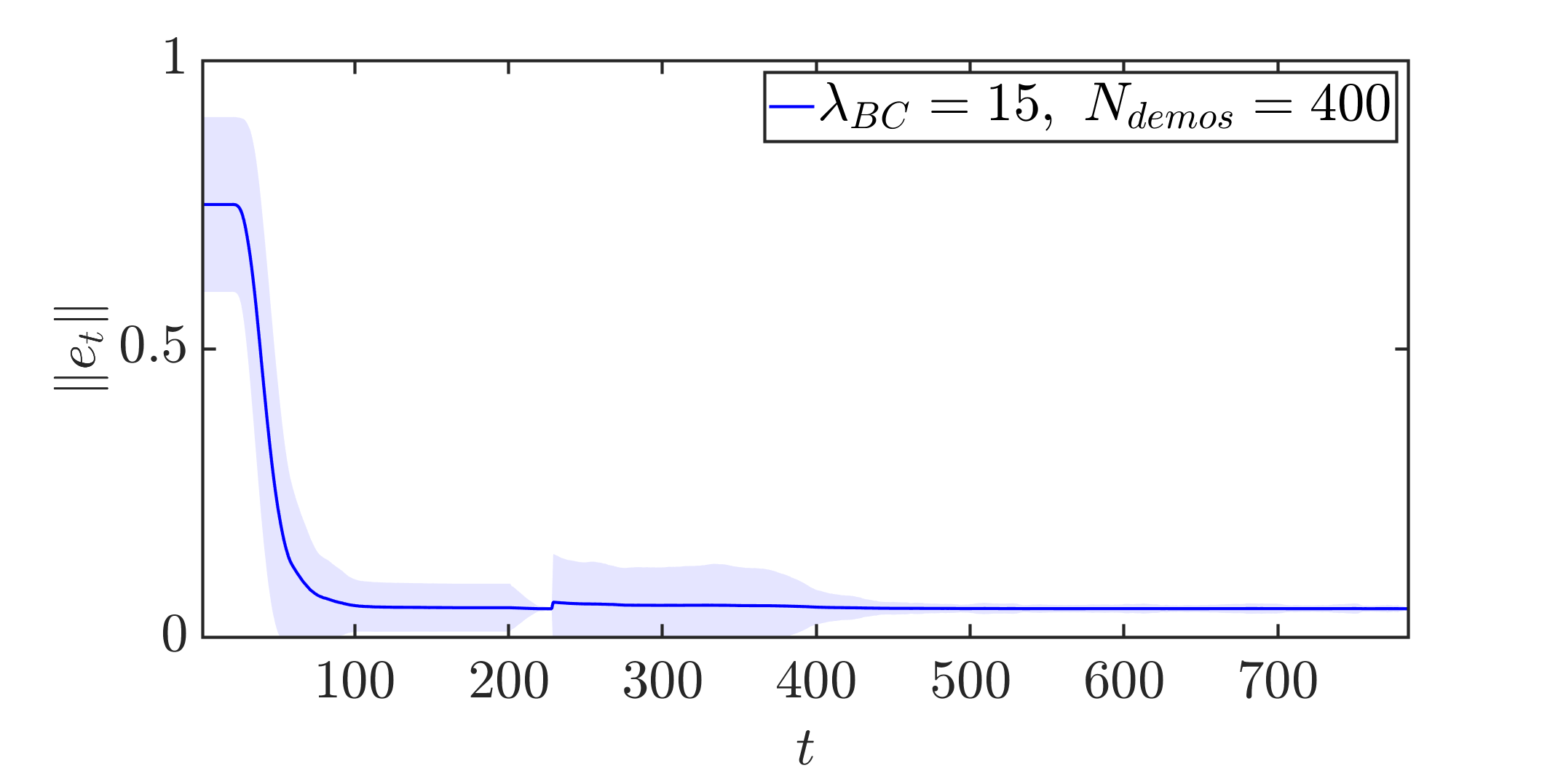}
    \end{subfigure}
    \begin{subfigure}[b]{0.48\textwidth}
        \includegraphics[width=0.98\textwidth,clip]{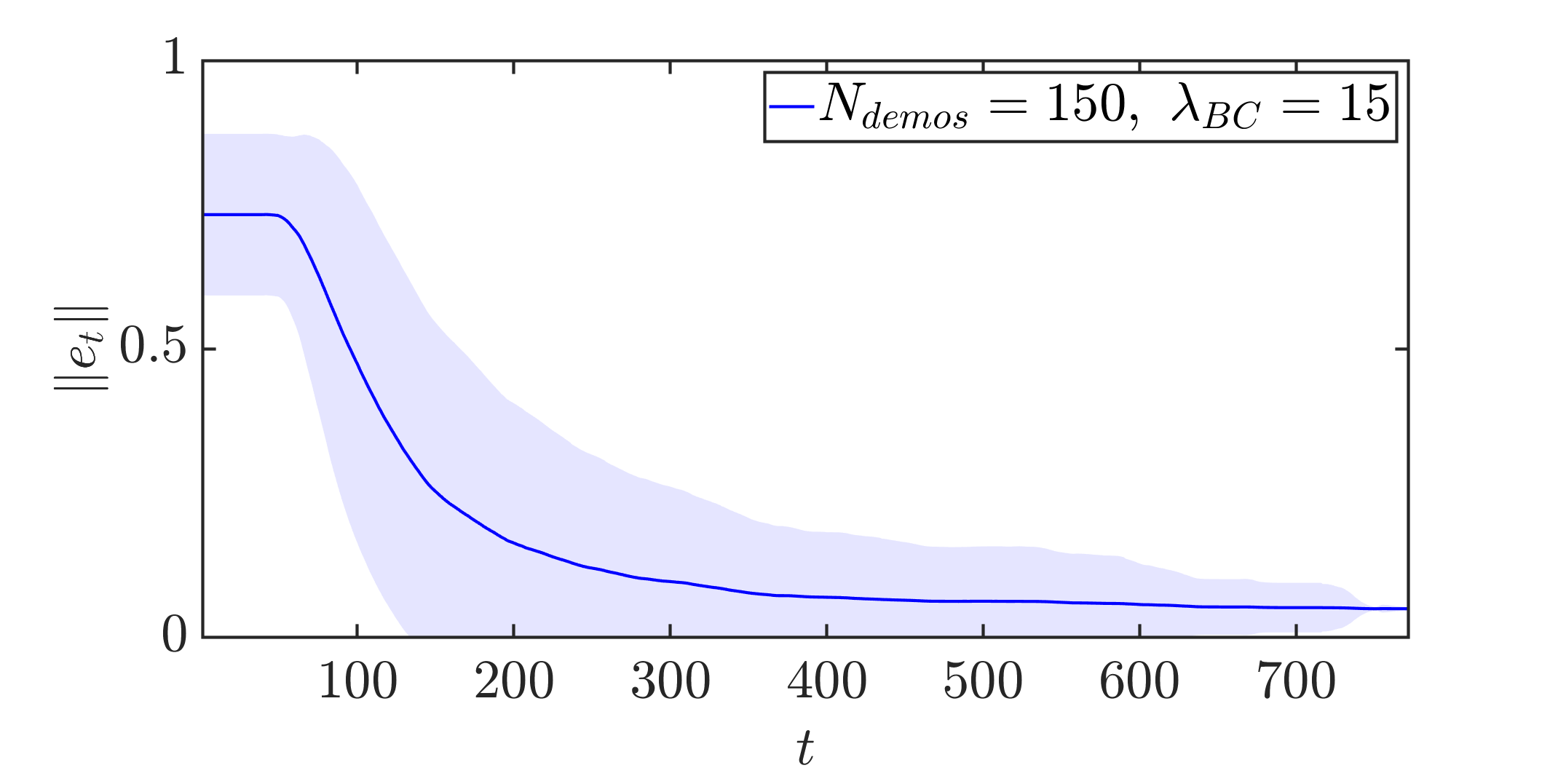}
    \end{subfigure}
    \begin{subfigure}[b]{0.48\textwidth}
        \includegraphics[width=0.98\textwidth,clip]{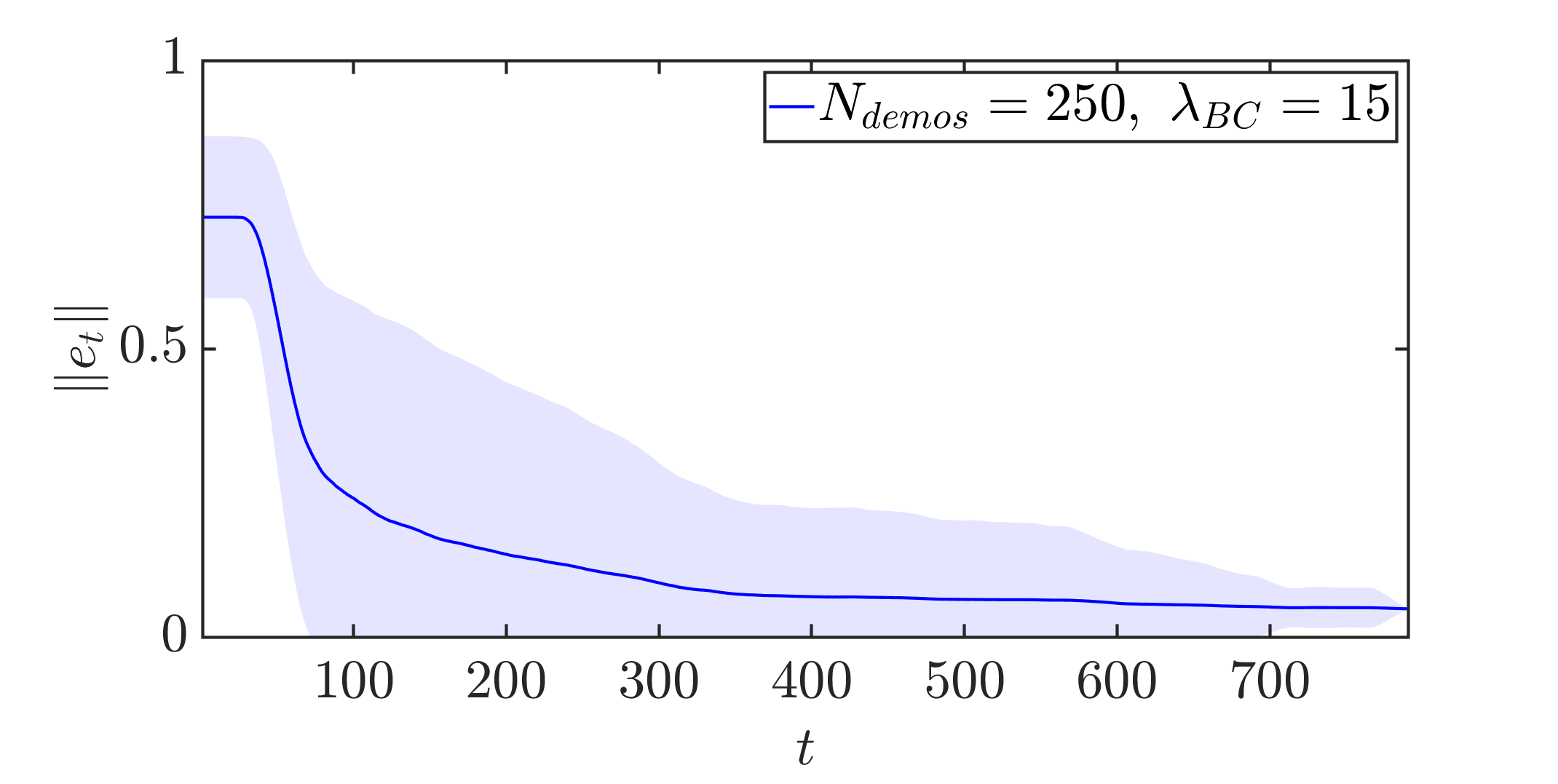}
    \end{subfigure}
    \begin{subfigure}[b]{0.48\textwidth}
        \includegraphics[width=0.98\textwidth,clip]{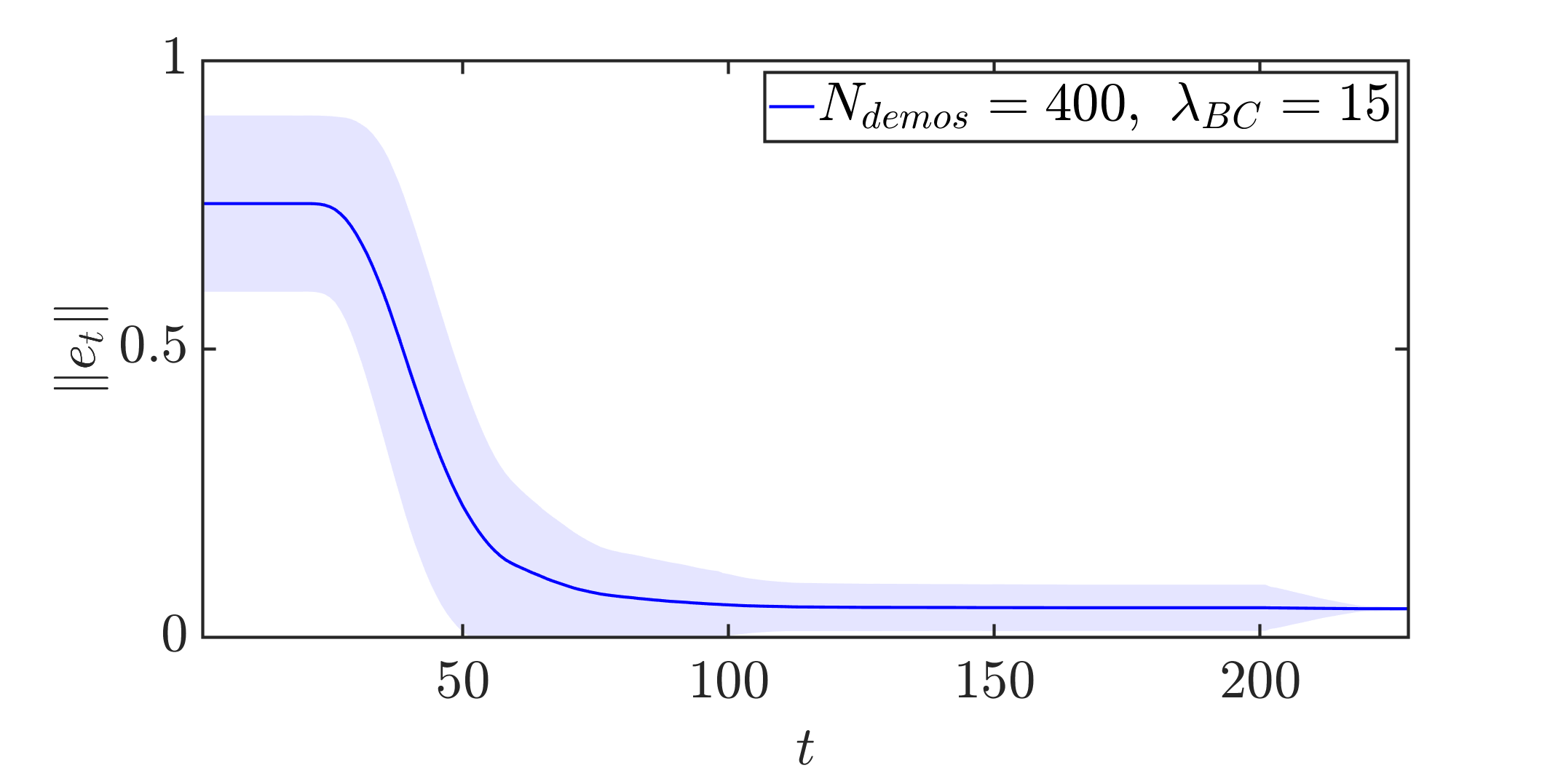}
    \end{subfigure}
    \caption{The average error $\|e_t\|$ in successful trials over time $t$ in the P\&P test.}
    \label{fig:pnp_errors}
\end{figure}

The reaching error trajectories $\|e_t\|$ of the P2P, P2P-O, and P\&P tasks in the test study, subject to different values of $\lambda_{BC}$ and $N_{\mathrm{demos}}$, are respectively illustrated in Figures \ref{fig:p2p_errors}, \ref{fig:p2po_errors}, and \ref{fig:pnp_errors}. In these figures, the dark curve represents the averaged trajectory over successful trials, and the shaded areas represent the standard deviation of the trajectories. It can be seen that implementing the behavior cloning and demonstrations generates smooth and stable trajectories in this test.

The numerical metrics of the P2P agent in the test study, subject to different hyperparameters $N_{\mathrm{demos}}$ and $\lambda_{BC}$, are presented in Table \ref{tab:p2p_imp}. Additionally, the deployment evaluation results of the tests for P2P-O and P\&P tasks are shown in Tables \ref{tab:p2po_imp} and \ref{tab:pnp_imp}. The PID controller designed as Equation \ref{eq:pid} is also tested for comparison. The results indicate that the PID baseline shows a mediocre success rate and ultimate error compared to the Demo-EASE agents. Meanwhile, PID also uses the lowest effort and gets the smallest reward. This confirms that PID is a conservative control baseline that produces smooth and energy-saving trajectories but neglects task-specific optimality. Note that the reaching errors are not relatively big because we truncate episodes at a $5\,$cm threshold.

\begin{table}[htbp]
\begin{center}
\begin{minipage}{0.8\linewidth}
\caption{Numerical evaluation results of the P2P agent in test}\label{tab:p2p_imp}
\begin{tabular*}{0.98\linewidth}{@{\extracolsep{\fill}}lccccccc@{\extracolsep{\fill}}}
\toprule%
\multirow{2}{*}{Agent} & \multicolumn{3}{@{}c@{}}{$\lambda_{BC}$ ($N_{\mathrm{demos}} = 100$)} & \multicolumn{3}{@{}c@{}}{$N_{\mathrm{demos}}$ ($\lambda_{BC}=1$)} & \multirow{2}{*}{PID} \\
\cmidrule{2-4}\cmidrule{5-7}%
& 0.1 & 0.6 & 1.8 & 0 & 80 & 160 & \\
\midrule
$P_{\mathrm{scs}} (\%)$   & 78.2       & 92.6       & 99.6       & 42.8       & 91.4       & 89.5   & 59.0      \\
$\overline{T}_{\mathrm{eff}}$         & 36.2         & 26.9         & 27.3         & 54.3         & 28.3         & 22.1   & 15.1        \\
$\overline{R}_{\mathrm{test}}$ & 9.88 & 9.96 & 9.98 & 9.97 & 9.98 & 9.94 & 9.58 \\
$\overline{E}_{\overline{95}}$ (cm) & 6.24 & 4.91 & 4.76 & 4.77 & 4.79 & 5.38 & 5.29 \\
\bottomrule
\end{tabular*}
\end{minipage}
\end{center}
\end{table}

\begin{table}[hbtp]
\begin{center}
\begin{minipage}{0.8\linewidth}
\caption{Numerical evaluation results of the P2P-O agent in test}\label{tab:p2po_imp}
\begin{tabular*}{0.98\linewidth}{@{\extracolsep{\fill}}lccccccc@{\extracolsep{\fill}}}
\toprule%
\multirow{2}{*}{Agent} & \multicolumn{3}{@{}c@{}}{$\lambda_{BC}$ ($N_{\mathrm{demos}} = 100$)} & \multicolumn{3}{@{}c@{}}{$N_{\mathrm{demos}}$ ($\lambda_{BC}=1$)} & \multirow{2}{*}{PID} \\\cmidrule{2-4}\cmidrule{5-7}%
& 0.5 & 1 & 2 & 100 & 200 & 400 & \\
\midrule
$P_{\mathrm{scs}} (\%)$                & 92.6       & 96.2       & 98.0       & 71.6       & 78.4       & 92.2   & 83.0      \\
$\overline{T}_{\mathrm{eff}}$                 & 67.3         & 66.1         & 65.8         & 70.5         & 67.6         & 66.1  & 54.9        \\
$\overline{R}_{\mathrm{test}}$ & 9.88 & 9.87 & 9.89 & 9.89 & 9.88 & 9.88 & 9.21 \\
$\overline{E}_{\overline{95}}$ (cm) & 5.43 & 5.30 & 5.15 & 5.06 & 5.28 & 5.27 & 5.51 \\
\bottomrule
\end{tabular*}
\end{minipage}
\end{center}
\end{table}

\begin{table}[hbtp]
\begin{center}
\begin{minipage}{0.8\linewidth}
\caption{Numerical evaluation results of the P\&P agent in test}\label{tab:pnp_imp}
\begin{tabular*}{0.98\linewidth}{@{\extracolsep{\fill}}lccccccc@{\extracolsep{\fill}}}
\toprule%
\multirow{2}{*}{Agent} & \multicolumn{3}{@{}c@{}}{$\lambda_{BC}$ ($N_{\mathrm{demos}} = 400$)} & \multicolumn{3}{@{}c@{}}{$N_{\mathrm{demos}}$ ($\lambda_{BC}=15$)} & \multirow{2}{*}{PID} \\
\cmidrule{2-4}\cmidrule{5-7}%
& 5 & 10 & 15 & 150 & 250 & 400 & \\
\midrule
$P_{\mathrm{scs}} (\%)$ & 80.8 & 90.0 & 97.6 & 86.2 & 89.2 & 97.6 & 73.4 \\
$\overline{T}_{\mathrm{eff}}$ & 56.0 & 45.8 & 43.5 & 44.1 & 44.7 & 43.5 & 44.0 \\
$\overline{R}_{\mathrm{test}}$ & 11.88 & 11.89 & 11.93 & 11.78 & 11.83 & 11.93 & 11.76 \\
$\overline{E}_{\overline{95}}$ (cm) & 5.29 & 4.86 & 5.09 & 6.07 & 4.86 & 5.09 & 6.05 \\
\bottomrule
\end{tabular*}
\end{minipage}
\end{center}
\end{table}

\subsection{Result Discussion}\label{sec:discussion}
We observed that the baseline RL agents stagnate in training due to difficult planning in a 4-DOF joint space. However, the agent begins learning by adding a relatively small number of demonstrations or applying behavior cloning guidance. It is also interesting to find that the behavior cloning gain contributes more to improving the training performance, compared to the number of demonstrations. In the test study, the Demo-agent also shows a decent performance in terms of safety (collision avoidance) and precision (small reaching error). Yet, trained RL agents applied more torque than PID, which can be regulated by reshaping the reward function.

It is worth mentioning that the GAS formulation does not require complete symmetry in the environment. While complete symmetry across an entire robotic workspace is uncommon in real-world scenarios, our framework does not require this condition. Demo-EASE is capable of leveraging local or partial symmetries, even when exogenous entities such as obstacles are present. For example, in the P2P-O environment used in our experiments, an obstacle introduces asymmetry in the workspace. However, by abstracting the space into subregions and mapping that obstacle into a symmetric partition, we are still able to replicate short demonstrations and boost the learning across the global environment. This suggests that the proposed framework is not limited to ideal symmetric environments but can be effectively applied in more complex or partially symmetric setups.
In practice, only some small regions of the environment are likely not to possess such symmetric properties. Nevertheless, our work is devoted to investigating the feasibility and potential of using demonstrations in symmetric environments to improve the sampling efficiency of RL methods with extreme use cases. The abstract demonstrations serve as generalized motion primitives that can be reused across structurally similar yet not identical regions, enabling improved bootstrapping and sample efficiency.

The simulation results presented here demonstrate the theoretical benefits and robustness of Demo-EASE. Extending this framework to real-world applications remains an important direction for future research. In particular, Demo-EASE could serve as a tool to bootstrap learning in physical robot deployments by leveraging a few safe, short demonstrations, where sample efficiency is critical. Additional future directions may include symmetry detection through geometric modeling of the environment, adaptations for symmetry-breaking entities, or incorporating human demonstrations.

Finally, while our current evaluation focuses on PPO and DDPG, these methods differ fundamentally in memory structure, actor stochasticity, and value function design. This makes them representative of a broader family of RL algorithms. The modular architecture of Demo-EASE, including the decoupled behavior cloning loss and the masking strategy, can be directly applied to other baselines such as SAC, TD3, TRPO, and A2C/A3C. Thus, additional experiments are not strictly necessary for validating generality, although they may further demonstrate extensibility.

\section{Concluding Remarks}\label{sec:conclusion}
In this paper, we have proposed a novel data aggregation method for model-free RL in robot manipulation and incorporated behavior cloning from abstract symmetric demonstration to improve both the training and test performance of the conventional RL. A conservative PID controller was used as the source of expert knowledge for producing such demonstrations. The symmetric property of the environment was exploited to duplicate the demonstration samples, which further improved the efficiency of the agent training.
The proposed dual-buffer structure in Demo-EASE causes only a minimal memory overhead since we intentionally constrain the number of demonstrations $N_{demos}$ to minimize the cloning influence. Compared to the substantial training burden of standard RL, which often requires a huge number of interactions to converge, the cost of memory usage and preprocessing for a small set of reusable demonstrations is negligible. Demo-EASE offset this with significantly reduced sample requirements and faster convergence.
In this work, we used a PID controller to generate demonstrations due to its simplicity and reliability, particularly in structured manipulation tasks. PID may not yield optimal policies; however, we aim to demonstrate that suboptimal controllers, implemented in small partitions, can provide useful guidance to improve RL sample efficiency. As a complementary direction, human demonstrations can also be used as a readily available source of data in real-world settings and support the broader goal of data-efficient RL without extensive pretraining.



\bibliography{bibliography}

\end{document}